%% file: main.tex
\documentclass[11pt, a4paper, logo, copyright]{deepmind}

\pdftrailerid{redacted}

\makeatletter
\renewcommand\bibentry[1]{\nocite{#1}{\frenchspacing\@nameuse{BR@r@#1\@extra@b@citeb}}}
\makeatother

\usepackage{kantlipsum, lipsum}
\usepackage{dsfont}
\usepackage{dm-colors}

\usepackage[authoryear, sort&compress, round]{natbib} 

\usepackage[utf8]{inputenc} 
\usepackage[T1]{fontenc}    
\usepackage{hyperref}       
\usepackage{url}            
\usepackage{booktabs}       
\usepackage{amsfonts}       
\usepackage{nicefrac}       
\usepackage{microtype}      
\usepackage{xcolor}         
\usepackage{graphicx}
\usepackage{siunitx}
\usepackage{caption}
\usepackage{color, colortbl}
\usepackage{subcaption}
\usepackage{wrapfig}
\usepackage{multirow}
\usepackage{xr}

\usepackage{etoolbox}
\let\classAND\AND
\let\AND\relax
\usepackage{algorithmic}

\let\AND\classAND
\AtBeginEnvironment{algorithmic}{\let\AND\algoAND}

\input{math_commands.tex}

\input{defns.tex}

\newcommand{\starcraft}{StarCraft~II}
\newcommand{\alphastar}{AlphaStar}
\newcommand{\scbenchmark}{AlphaStar Unplugged}
\newcommand{\veryhard}{\texttt{very\_hard}}

\newcommand{\mz}{{MuZero}}
\newcommand{\mzsup}{{MuZero Supervised}}
\newcommand{\mzsa}{{Sampled MuZero}}

\newcommand{\mzun}{{MuZero Unplugged}}

\title{\scbenchmark{}: Large-Scale Offline Reinforcement Learning}

\correspondingauthor{mmathieu@google.com}

\keywords{Starcraft II, Offline RL, Large-scale learning} 

\author[*,1]{Micha\"el~Mathieu}
\author[*,1]{Sherjil~Ozair}
\author[*,1]{Srivatsan~Srinivasan}
\author[*,1]{Caglar~Gulcehre}
\author[*,2]{Shangtong~Zhang}
\author[*,1]{Ray~Jiang}
\author[*,1]{Tom~Le~Paine}
\author[1]{Richard~Powell}
\author[1]{Konrad~\.Zo\l{}na}
\author[1]{Julian~Schrittwieser}
\author[1]{David~Choi}
\author[1]{Petko~Georgiev}
\author[1]{Daniel~Toyama}
\author[1]{Aja~Huang}
\author[1]{Roman~Ring}
\author[1]{Igor~Babuschkin}
\author[1]{Timo Ewalds}
\author[1]{Mahyar Bordbar}
\author[1]{Sarah~Henderson}
\author[1]{Sergio~G\'omez~Colmenarejo}
\author[1]{A\"aron~van~den~Oord}
\author[1]{Wojciech~Marian~Czarnecki}
\author[1]{Nando de Freitas}
\author[1]{Oriol Vinyals}

\affil[*]{Equal contributions}
\affil[1]{Google DeepMind}
\affil[2]{University of Virgina}

\begin{abstract}
\starcraft{} is one of the most challenging simulated reinforcement learning environments; it is partially observable, stochastic, multi-agent, and mastering \starcraft{} requires strategic planning over long time horizons with real-time low-level execution. It also has an active professional competitive scene. \starcraft{} is uniquely suited for advancing offline RL algorithms, both because of its challenging nature and because Blizzard has released a massive dataset of millions of \starcraft{} games played by human players. This paper leverages that and establishes a benchmark, called \emph{\scbenchmark{}}, introducing unprecedented challenges for offline reinforcement learning. We define a dataset (a subset of Blizzard's release), tools standardizing an API for machine learning methods, and an evaluation protocol. We also present baseline agents, including behavior cloning, offline variants of actor-critic and \mz{}. We improve the state of the art of agents using only offline data, and we achieve 90\% win rate against previously published \alphastar{} behavior cloning agent. 
\end{abstract}

\begin{document}
\maketitle

\section{Introduction}
\label{sec:introduction}

Deep Reinforcement Learning is dominated by online Reinforcement Learning (RL) algorithms, where agents must interact with the environment to explore and learn.
The online RL paradigm achieved considerable success on Atari \citep{mnih2015human}, Go \citep{silver2017mastering}, \starcraft{} \citep{vinyals2019grandmaster}, DOTA 2 \citep{berner2019dota}, and robotics \citep{andrychowicz2020learning}.
However, the requirements of extensive interaction and exploration make these algorithms unsuitable and unsafe for many real-world applications. In contrast, in the offline setting \citep{fu2020d4rl, fujimoto2019off,gulcehre2020rl}, agents learn from a fixed dataset previously logged by humans or other agents. While the offline setting would enable RL in real-world applications, most offline RL benchmarks such as D4RL \citep{fu2020d4rl} and RL Unplugged \citep{gulcehre2020rl} have mostly focused on simple environments with data produced by RL agents. More challenging benchmarks are needed to make progress towards more ambitious real-world applications.

To rise to this challenge, we introduce \emph{\scbenchmark{}}, an offline RL benchmark, which uses a dataset derived from replays of millions of humans playing the multi-player competitive game of \starcraft{}. \starcraft{} continues to be one of the most complex simulated environments, with partial observability, stochasticity, large action and observation spaces, delayed rewards, and multi-agent dynamics. Additionally, mastering the game requires strategic planning over long time horizons, and real-time low-level execution. Given these difficulties, breakthroughs in \scbenchmark{} will likely translate to many other offline RL settings, potentially transforming the field.

Additionally, unlike most RL domains, \starcraft{} has an independent leaderboard of competitive human players over a wide range of skills. It constitutes a rich and abundant source of data to train and evaluate offline RL agents.



With this paper, we release the most challenging large-scale offline RL benchmark to date, including the code of canonical agents and data processing software. We note that removing the environment interactions from the training loop significantly lowers the compute demands of \starcraft{}, making this environment accessible to far more researchers in the AI community.

Our experiments on this benchmark suggest that families of algorithms that are state-of-the-art on small scale benchmarks do not perform well here, e.g. Return Conditioned Behavior Cloning \citep{srivastava2019training, emmons2021rvs}, Q-function based approaches \citep{fujimoto2019off, wang2020critic, gulcehre2021regularized}, and algorithms that perform off-policy evaluation during learning \citep{schrittwieser2021online}. These approaches sometimes fail to win a single game against our weakest opponent, and all fail to outperform our unconditional behavior cloning baseline.

However, it has also provided insights on how to design successful agents. So far all of our successful approaches are so-call one-step offline RL approaches \citep{gulcehre2021regularized, brandfonbrener2021offline}. Generally, our best performing agents follow a two-step recipe: first train a model to estimate the behavior policy and behavior value function. Then, use the behavior value function to improve the policy, either while training or during inference. We believe sharing these insights will be valuable to anyone interested in offline RL, especially at large scale.


\section{\starcraft{} for Offline Reinforcement Learning}
\label{sec:sc2_orl}

StarCraft is a real-time strategy game in which players compete to control a shared map by gathering resources and building units and structures. The game has several modes, such as team games or custom maps. For instance, the StarCraft Multi-Agent Challenge \citep{DBLP:journals/corr/abs-1902-04043} is an increasingly popular benchmark for Multi-Agent Reinforcement Learning and includes a collection of specific tasks.

In this paper, we consider \starcraft{} as a two-player game, which is the primary setting for \starcraft{}. This mode is played at all levels, from casual online games to professional esport. It combines high-level reasoning over long horizons with fast-paced unit management. There are numerous strategies for \starcraft{} with challenging properties presenting cycles and non-transitivity, especially since players start the game by selecting one of three alien \emph{races}, each having fundamentally different mechanics, strengths and weaknesses. Each game is played on one of the several \emph{maps}, which have different terrain and can affect strategies.

\starcraft{} has many properties making it a great environment to develop and benchmark offline reinforcement learning algorithms. It has been played online for many years, and millions of the games were recorded as \emph{replays}, which can be used to train agents. 
On the other hand, evaluation of the agents can be done by playing against humans --- including professional players --- the built-in bots, scripted bots from the community, or even the stronger online RL agents such as AlphaStar \citep{vinyals2019grandmaster} or TStarBot \citep{han2021tstarbotx}.
Finally, we highlight a few properties of \starcraft{} that make it particularly challenging from an offline RL perspective.

\textbf{Action space.} When learning from offline data, the performance of algorithms depends greatly on the availability of different state-action pairs in the data. We call this \emph{coverage} --- the more state-action pairs are absent, \emph{i.e.} the lower the coverage, the more challenging the problem is. \starcraft{} has a highly structured action space. The agent must select an action type, select a subset of its units to apply the action to, select a target for the action (either a map location or a visible unit), and decide when to observe and act next. In our API, we can consider there are approximately $10^{26}$ possible actions per game step. In comparison, Atari has only $18$ possible actions per step. This makes it almost impossible to attain high state-action coverage for \starcraft{}.

\textbf{Stochastic environment.} Stochastic environments may need many more trajectories to obtain high state-action coverage. The game engine has a small amount of stochasticity itself, but the main source of randomness is the unknown opponent policy, which is typically not deterministic. In conrast, in the Atari environment, stochasticity arises only from sticky actions \citep{machado2018revisiting}.

\textbf{Partial Observability.} \starcraft{} is an imperfect information game. Players only have information about opponent units that are within the field of view of the player's own units.
As a result, players need to scout, \emph{i.e.}\ send their units around the map to gather information about the current state of the game, and may need it at a later point in the game. On the other hand, a memory of the 3 previous frames is usually considered sufficient for Atari.

\textbf{Data.} For \starcraft{}, we have access to a dataset of millions of human replays. These replays display a wide and diverse range of exploration and exploitation strategies. In comparison, the existing benchmarks \citep{gulcehre2020rl,agarwal2020optimistic} have a bias toward datasets generated by RL agents.

\section{\scbenchmark{}}
\label{sec:alphastar_unplugged}

We propose \scbenchmark{} as a benchmark for offline learning on \starcraft{}. This work builds on top of the \starcraft{} Learning Environment and associated replay dataset \citep{vinyals2017starcraft}, and the AlphaStar agents described in \cite{vinyals2019grandmaster}, by providing a few key components necessary for an offline RL benchmark:
\begin{itemize}
    \item \textbf{Training setup.} We fix a dataset and a set of rules for training in order to have fair comparison between methods. 
    \item \textbf{Evaluation metric.} We propose a set of metrics to measure performance of agents.
    \item \textbf{Baseline agents.} We provide a number of well tuned baseline agents.
    \item \textbf{Open source code.} Building an agent that performs well on \starcraft{} is a massive engineering endeavor. We provide a well-tuned behavior cloning agent which forms the backbone for all agents presented in this paper\footnote{We open-sourced our architecture, data pipeline, dataset generation scripts and supervised learning agent in \url{https://github.com/deepmind/alphastar}}.
\end{itemize}

\subsection{Dataset}
\label{section:dataset}

\begin{wrapfigure}{r}{0.36\textwidth}
    \centering
     \includegraphics[width=\linewidth]{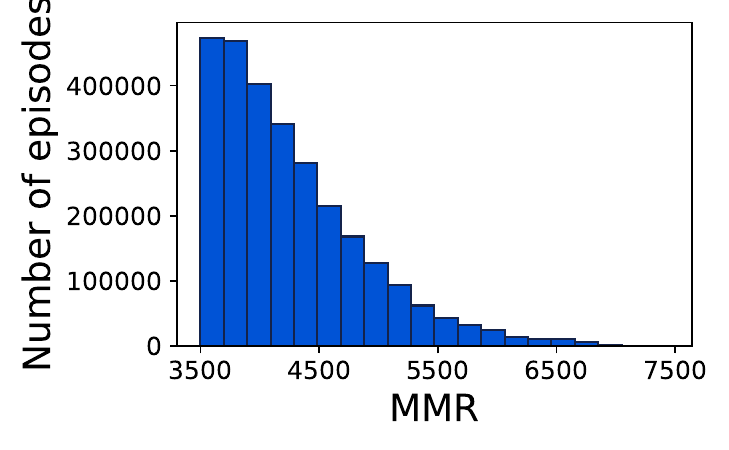}
    \caption{Histogram of player MMR from replays used for training.}
    \label{fig:mmr}
\end{wrapfigure}

About 20 million \starcraft{} games are publicly available through the replay packs\footnote{\url{https://github.com/Blizzard/s2client-proto/tree/master/samples/replay-api}}. For technical reasons, we restrict the data to \starcraft{} versions 4.8.2 to 4.9.2 which leaves nearly 5 million games. They come from the \starcraft{} \emph{ladder}, the official matchmaking mechanism. Each player is rated by their \emph{MMR}, a ranking mechanism similar to \emph{Elo} \citep{elo1978rating}. The MMR ranges roughly from 0 to 7000. Figure~\ref{fig:mmr} shows the distribution of MMR among the episodes. In order to get quality training data, we only use games played by players with MMR greater than 3500, which corresponds to the top 22\% of players. This leaves us with approximately 1.4 million games. Each game forms two episodes from a machine learning point of view --- one for each side, since we consider two-player games --- so there are 2.8 million episodes in the dataset. This represents a total of more than 30 years of game played. These replays span over two different balance patches, introducing some subtle differences in the rules of \starcraft{} between the older and the more recent games, which are small enough to be ignored\footnote{However, the version of the game is available for each episode, so one could decide to condition the agent on the version.}. In addition, the map pool changed once during this period, so the games are played on a total of 10 different maps\footnote{Acropolis, Automaton,  Cyber Forest, Kairos Junction, King's Cove, New Repugnancy, Port Aleksander, Thunderbird, Turbo Cruise '84, Year Zero.}.

The average two-player game is about $11$ minutes long which corresponds to approximately $15,000$ internal game steps in total. This poses a significant modeling challenge, making training harder and slower. Therefore, we shorten trajectories by only observing the steps when the player took an action. We augment each observation by adding the \emph{delay} which contains the number of internal game steps until the next action, and we discard the internal steps in-between. This cuts the effective length of the episode by $12$ times, and is similar to what was done in \cite{vinyals2019grandmaster}.

Each episode also contains metadata, the most important ones being the outcome, which can be $1$ for a victory, $0$ for a draw\footnote{Draws are rare in \starcraft, but can happen if no player can fulfill the win-condition.} and $-1$ for a defeat, as well as the MMR of each player. The games were played online using Blizzard's matchmaking system which ensures that in the vast majority of games, both players have a similar MMR.

The replays are provided by Blizzard and hosted on their servers. The data is anonymized, and does not contain personal information about the players. The full dataset represents over 30 years of game play time, in the form of 21 billion internal game steps. This corresponds to 3.5 billion training observations.


\subsection{Training restrictions}
\label{sec:training}

\begin{wrapfigure}{r}{0.33\textwidth}
\vspace{-0.5cm}
    \centering
        \includegraphics[width=1\linewidth]{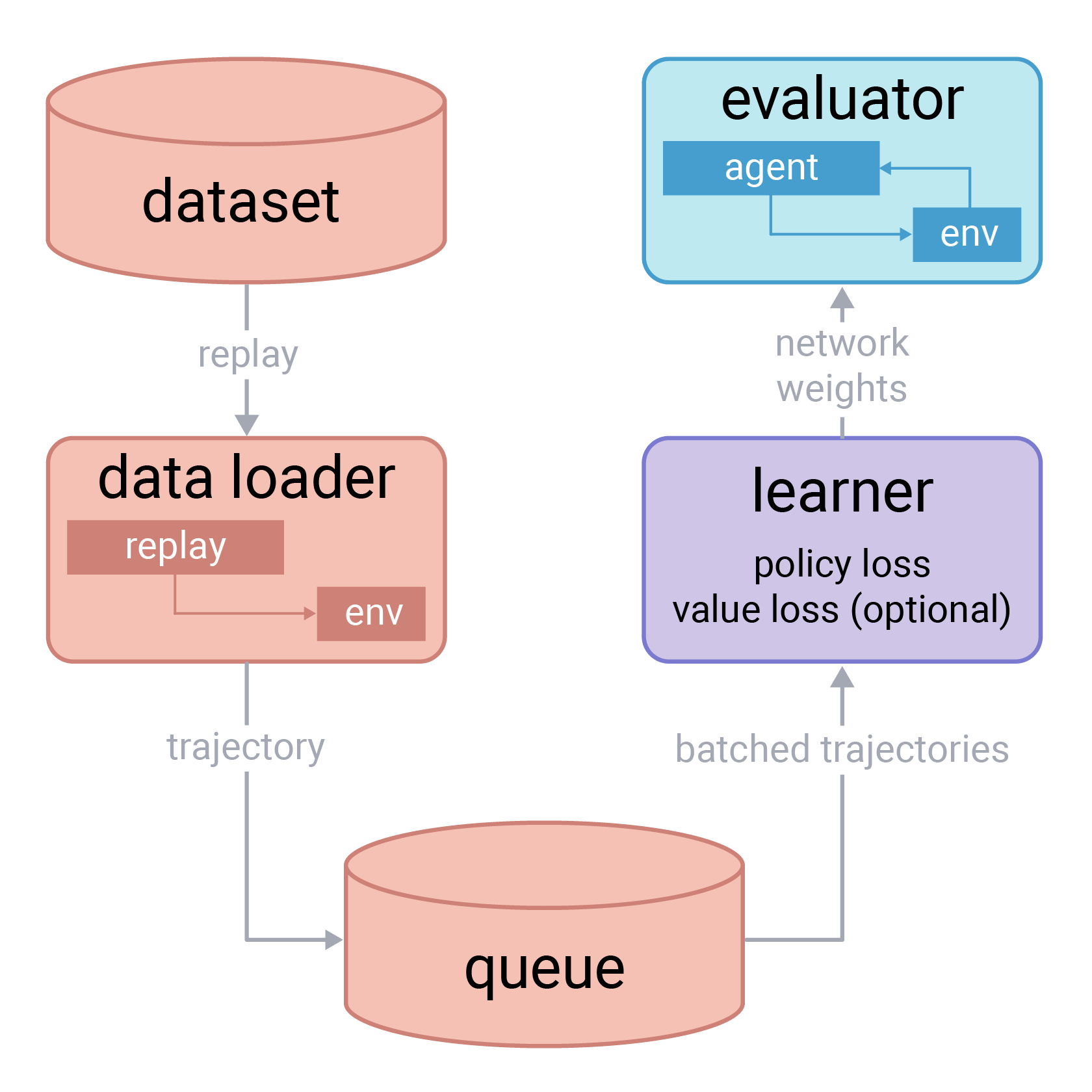}
    \caption{Training procedure.}
    \label{fig:training}
\vspace{-0.3cm}
\end{wrapfigure}

During training, we do not allow algorithms to use data beyond the dataset described in Section~\ref{section:dataset}. In particular, the environment cannot be used to collect more data. However, online policy evaluation is authorized, \emph{i.e.} policies can be run in the environment to measure their performance. This may be useful for hyperparameter tuning.

Unlike the original AlphaStar agents, agents are trained to play all three races of \starcraft{}. This is more challenging, as agents are typically better when they are trained on a single race. They are also trained to play on all 10 maps available in the dataset.

In our experiments, we tried to use the same number of training inputs whenever possible --- of the order of $k_{max}=10^{10}$ observations in total --- to make results easier to compare. However this should be used as a guideline and not as a hard constraint. The final performance reached after each method eventually saturates is a meaningful comparison metric, assuming each method was given enough compute budget.

\subsection{Evaluation protocol}
\label{sec:evaluation}

Numerous metrics can be used to evaluate the agents. On one hand, the easiest to compute --- and least informative --- is simply to look at the value of the loss function. On the other hand, perhaps the most informative --- and most difficult to compute --- metric would be to evaluate the agent against a wide panel of human players, including professional players. In this paper, we propose a compromise between these two extremes. We evaluate our agents by playing repeated games against a fixed selection of 7 opponents: the \veryhard{} built-in bot\footnote{The \veryhard{} bot is not the strongest built-in bot in \starcraft, but it is the strongest whose strength does not come from unfair advantages which break the game rules.}, as well as a set of 6 reference agents presented below. 

During training, we only evaluate the agents against the \veryhard{} bot, since it is  significantly less expensive, and we mostly use that as a validation metric, to tune hyper-parameters and discard non-promising experiments. 

Fully trained agents are evaluated against the full set of opponents presented above, on all maps. We combine these win rates into two aggregated metrics while uniformly sampling the races of any pair of these agents:  \emph{Elo rating} \citep{elo1978rating}, and \emph{robustness}. Robustness is computed as one minus the minimum win rate over the set of reference agents. See details of the metrics computation in Appendix~\ref{app:eval}.

\section{Reference Agents}

As explained in Section~\ref{sec:alphastar_unplugged}, we provide a set of 6 reference agents, which can be used both as baselines and for evaluation metrics. In this section, we detail the methodology and algorithms used to train them. The implementation details can be found in Appendix~\ref{app:implementations}, and results in Section~\ref{sec:reference_agents_evaluation}.

\subsection{Definitions}

The underlying system dynamics of \starcraft{} can be described by a \emph{Markov Decision Process}\footnote{Strictly speaking, we have a Partially Observable MDP, but we simplify this for ease of presentation.} \emph{(MDP)} \citep{bellman1957}.
An MDP, $(\gS, \gA, \trans, r, \gI)$, consists of finite sets of states $\gS$ and actions $\gA$, a transition distribution $\trans(s'|s,a)$ for all $(s,a,s')\in\gS\times\gA\times\gS$, a reward function\footnote{In the usual definition of an MDP, the reward is a function of the state and the action. But in \starcraft{}, the reward is 1 in a winning state, -1 in a losing state, and zero otherwise. So it does not depend on the action.} $r:\gS \rightarrow \mathbb{R}$, and an initial state distribution $\gI: \gS \rightarrow [0,1]$. In the offline setting, the agent does not interact with the MDP but learns only from a dataset $\data$ containing \emph{episodes}, made of sequences of state and actions $\left(s_t, a_t\right)$. We denote $\mathbf{s}$ the sequence of all states in the episode, and $len(\mathbf{s})$ its length. A \emph{policy} is a probability distribution over the actions given a state. The dataset $\data$ is assumed to have been generated by following an unknown {\it behavior policy} $\mu$, such that $a_t\sim\mu(\cdot|s_t)$ for all $t<len(\mathbf{s})$.

As explained in Section~\ref{section:dataset}, observed states are a subset of the internal game steps. We call \emph{delay} the number of internal game steps between two observed internal game steps, which corresponds to the amount of real time elapsed between the two observations\footnote{One internal game step occurs every 45ms.}. Given states and action $(s_t, a_t, s_{t+1})$, we note $d(a_t)$ the delay between states $s_t$ and $s_{t+1}$. Note that the delay at step $t$ is referring to the step $t+1$, not $t-1$. This is needed for inference, since the environment must be provided with the number of internal steps to skip until the next observation. Therefore the delay must be part of the action.

Given a policy $\pi$ and a state $s_t$, we define the expected discounted return $v^\pi(s_t)$ as the expected sum of the discounted rewards obtained if we follow $\pi$ from $s_t$. The discount between two steps is based on the delay between the steps. In the case of \starcraft{}, the reward is the win-loss signal, so it can only be non-zero on the last step of the episode. Therefore we can write
\begin{align}
\label{eq:goal}
v^{\pi}(s_t) &= \mathbb{E}_{s_{k+1} \sim{}(\cdot \mid s_k, a_k), a_k \sim \pi{}(\cdot \mid s_k), \forall k\geq{}t} \left[\gamma^{D_t(\mathbf{s})} r(\mathbf{s})\right] \qquad \text{with}\qquad D_t(\mathbf{s})=\sum_{k=t}^{len(\mathbf{s})-1}d(a_k),
\end{align}
where $r(\mathbf{s})$ is the reward on the last step of the episode. $D_t(\mathbf{s})$ is simply the remaining number of internal game steps until the end of the episode.

The goal of offline RL is to find a policy $\pi^*$ which maximizes $\mathbb{E}_{s_0\in\gI}[v^{\pi^*}(s_0)]$. Dunring training, we refer to the policy $\pi$ trained to estimate $\pi^*$ as the {\it target policy}.

We use $V^{\mu}$ and $V^{\pi}$ to denote the value functions for the behavior and target policies $\mu$ and $\pi$, which are trained to estimate $v^\mu$ and $v^\pi$, respectively.

We typically train the agent on \emph{rollouts}, \emph{i.e.} sequences of up to $K$ consecutive timesteps, assembled in a minibatch of $M$ independent rollouts. Unless specified otherwise, the minibatches are independent from each other, such that two consecutive minibatches are not correlated.



\subsection{Architecture}\label{sec:architecture}

\begin{figure}
    \centering
    \includegraphics[width=\textwidth]{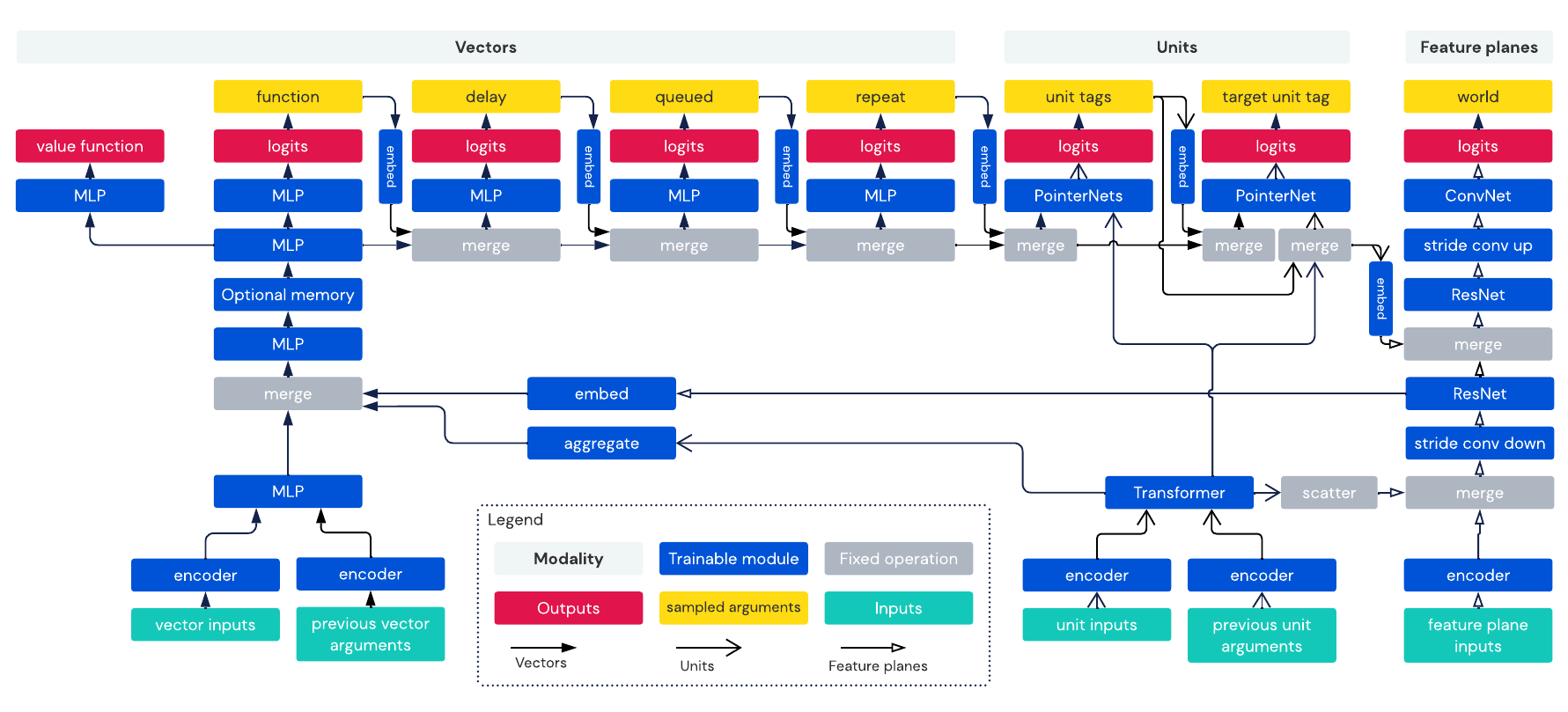}
    \caption{Illustration of the architecture that we used for our reference agents. Different types of data are denoted by different types of arrows (vectors, units or feature planes).}
    \label{fig:architecture}
\end{figure}

All our experiments are based on the same agent architecture. It is an improved version of the model used in \cite{vinyals2019grandmaster}. The full architecture is summarized on Figure~\ref{fig:architecture}.

Inputs of the raw \starcraft{} API are structured around three modalities: \emph{vectors}, \emph{units} --- a list of features for each unit present in the game --- and \emph{feature planes} (see Appendix~\ref{app:sc2_interface} for more details).

Actions are comprised of seven \emph{arguments}, and can be organized in similar modalities: \texttt{function}, \texttt{delay}, \texttt{queued} and \texttt{repeat} as vectors, since each argument is sampled from a single vector of logits. \texttt{Unit\_tags} and \texttt{target\_unit\_tag} refer to indices in the units inputs. Finally, the \texttt{world} action is a 2d point on the feature planes.

We structure the architecture around these modalities:
\begin{itemize}
    \item Each of the three modalities of inputs is encoded and processed independently using a fitting architecture: MLP for the vector inputs, transformer \citep{transformers} for the units input and residual convolutional network \citep{he2015deep} for the feature planes. Some of these convolutions are strided so that most of the computation is done at a lower resolution. Arguments of the previous action are embedded as well, with the execption of the \texttt{world} previous argument, since we found this causes too much overfitting.
    \item We use special operations to add interactions between these modalities: we \emph{scatter} units into feature planes, \emph{i.e.} we place the embedding of each unit in its corresponding spatial location on the feature plane. We use a averaging operation to embed the units into the embedded vectors. Feature planes are embedded into vectors using strided convolutions and reshaping, and the reverse operations to embed vectors into feature planes.
    \item We tried using memory in the vector modality, which can be LSTM \citep{lstm} or Transformer XL \citep{transformerXL}. Most of our results do not use memory (see Section~\ref{sec:memory}).
    \item For the experiments using a value function, we add an MLP on top of the vector features to produce a estimate of the value function.
    \item Finally, we sample actions. The seven arguments are sampled in the following order: \texttt{function}, \texttt{delay}, \texttt{queued}, \texttt{repeat}, \texttt{unit\_tags}, \texttt{target\_unit\_tag} and \texttt{world}. They are sampled autoregressively,\hspace{-.06em}\footnote{With the exception of \texttt{target\_unit\_tag} and \texttt{world}, because no action in the API uses a \texttt{target\_unit\_tag} and a \texttt{world} argument at the same time.} \emph{i.e.} each sampled argument is embedded to sample the next one. The first four arguments are sampled from the vector modality. The next two are sampled from the vector and units modalities using pointer networks \citep{vinyals2017pointer}, and finally the \texttt{world} argument is sampled from the upsampled feature planes. Note that \texttt{unit\_tags} is actually obtained by sampling the pointer network 64 times autoregressively, so conceptually, \texttt{unit\_tags} represent 64 arguments.
\end{itemize}

The exact hyperparameters and details of the architecture can be found in the open-sourced code which can be accessed via \url{https://github.com/deepmind/alphastar}.

\paragraph{MMR conditioning.}
At training time, the MMR of the player who generated the trajectory is passed as a vector input. During inference, we can control the quality of the game played by the agent by changing the MMR input. In practice, we set the MMR to the highest value to ensure the agent plays its best. This is similar to Return-Conditioned Behavior Cloning \citep{srivastava2019training} with the MMR as the reward.

\begin{figure}
    \centering
    \includegraphics[width=\textwidth]{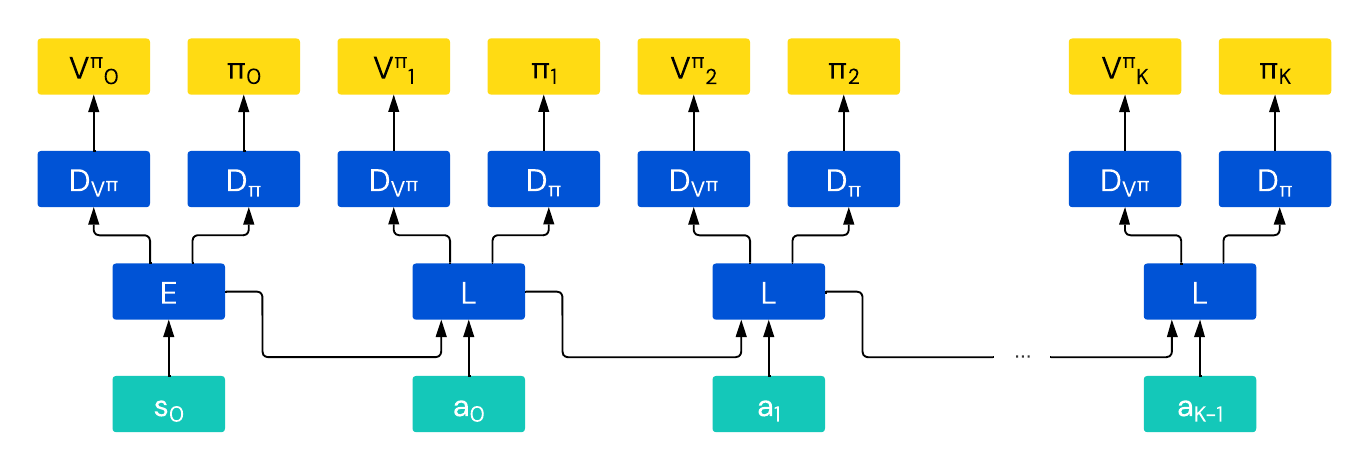}
    \caption{Illustration of the architecture used for MuZero. $E$ is the encoder, $L$ is the latent model, and $D_\pi$ and $D_{V^\pi}$ are the policy and value function decoders, respectively.}
    \label{fig:muzero-architecture}
\end{figure}

\paragraph{MuZero latent model.}
For the MuZero experiments, detailed in Section~\ref{sec:muzero}, we define the latent space $\gL$ as the space of vectors before the \texttt{function} MLP. We split the model presented above into an encoder $E : \gS\to\gL$ and two decoder: $D_\pi$ maps latent states to distributions over actions, and a value function decoder $D_{V^\pi} : \gL\to\sR$, such that $\pi(\cdot|s)=D_\pi(E(s))$ and $V^\pi(s)=D_{V^\pi}(E(s))$. Note that in our implementation, the decoder $D_\pi$ actually produces distributions for the \texttt{function} and \texttt{delay} only. The other arguments are obtained from the estimated behavior policy $\hat\mu$. Finally, we add a latent model $L:\gL\times\gA\to \gL$. Given a rollout $((s_0, a_0), ... (s_K, a_K))$, we compute:
\begin{align}
\label{eq:muzero-model}
    h_0 &= E(s_0) & h_{k+1} &= L(h_k, a_k) & \pi(\cdot|s_k) &= D_\pi(h_k) & V^\pi(s_k) &= D_{V^\pi}(h_k)
\end{align}
for all $k < K$. Note that $s_0$ is only the first state of the rollout, but not necessarily the first state of an episode. See Figure~\ref{fig:muzero-architecture} for an illustration.


\subsection{Behavior cloning}
\label{sec:bc}

\paragraph{Behavior Cloning (BC) agent.}
Our first reference agent is trained using \emph{behavior cloning}, the process of estimating the behavior policy $\mu$. We learned an estimate $\hat{\mu}$ by minimizing the negative log-likelihood of the action $a_t$ under the policy $\hat{\mu}(\cdot|s_t)$. Given a rollout $\mathbf{s}$, we write
\begin{align}
    L^{BC}(\mathbf{s}) = -\sum_{t=0}^{len(\mathbf{s})-1}\log\left(\hat{\mu}(a_t|s_t)\right) .
\end{align}
This is similar to training a language model. The procedure is detailed in Algorithm~\ref{alg:bc} in the Appendix. It is the same procedure that was used by the AlphaStar Supervised agent in \cite{vinyals2019grandmaster}. In practice, since each action is comprised of seven arguments, there is one loss per argument. \\
In order to avoid overfitting during behavior cloning, we also used a weight decay loss which is defined as the sum of the square of the network parameters.

\paragraph{Fine-tuned Behavior Cloning (FT-BC) agent.}
Behavior Cloning mimics the training data, so higher quality data should lead to better performance. Unfortunately, since filtering the data also decreases the number of episodes, generalization is affected (see Section~\ref{sec:dataset_filtering}).
In order to get the best of both worlds, we used a method called \emph{fine tuning}. It is a secondary training phase after running behavior cloning on the whole dataset. In this phase, we reduced the learning rate and filter the data to top-tier games. This generalizes better than training only on either set of data, and was already used in \cite{vinyals2019grandmaster}.

\subsection{Offline Actor-Critic}
\label{sec:oac}

Actor-critic \citep{witten1977adaptive, barto1983neuronlike} algorithms learn a target policy $\pi$ and the value function $V^\pi$. In off-policy settings, where the target policy $\pi$ differs from the behavior policy $\mu$, we compute \emph{importance sampling} ratios $\rho_t(a_t|s_t) = \pi(a_t|s_t)/\mu(a_t|s_t)$ where $(s_t, a_t)$ come from the data, \emph{i.e.} follow the behavior policy. There are many variants of the loss in the literature. The simplest version is called 1-Step Temporal Differences, or TD(0), defined as:
\begin{align}
\label{eq:td-zero}
    L^{TD(0)}(\mathbf{s}) = -\sum_{t=0}^{len(\mathbf{s})-2}\mathbf{\circleddash}\!\left[\rho_t(a_t|s_t)\left(\gamma V^{\pi}(s_{t+1}) - V^{\pi}(s_t) + r(s_{t+1})\right)\right]\log(\pi(a_t|s_t))
\end{align}
where $\mathbf{\circleddash}$ symbol corresponds to the stop-gradient operation. In this equation, $V^\pi$ is called the \emph{critic}. The loss can be modified to use N-Step Temporal Differences \citep{sutton2018} by adding more terms to Equation~\ref{eq:td-zero}, and it can be further improved by using V-Trace \citep{impala} in order to reduce variance. Note that for simplicity of implementation, we only applied this loss for some of the arguments, namely \texttt{function} and \texttt{delay}, and we use the behavior policy for the other ones.

We learned the estimated behavior value $V^\mu$ by minimizing the \emph{Mean-Squared Error (MSE)} loss:
\begin{align}
    L^{MSE}(\mathbf{s}) = \frac{1}{2}\sum_{t=0}^{len(\mathbf{s})-1}||V^\mu(s_t) - r(\mathbf{s})||_2^2 \ .
\end{align}

\paragraph{Offline Actor-Critic (OAC) agent.}
Although actor-critic has an off-policy correction term, it was not enough to make it work without adjustments to the pure offline setting.

The behavior policy $\mu$ appears in the denominator of $\rho$, but we do not have access to the behavior policy used by the players, we can only observe their actions. Fortunately, the behavior Cloning agent learns an estimate $\hat\mu$ which we used to compute the estimated $\hat\rho = \pi/\hat\mu$.

The Behavior Cloning policy $\hat\mu$ can be used as the starting point for $\pi$  (\emph{i.e.}\ used to initialize the weights). This way, the estimated importance sampling $\hat\rho$ equals $1$ at the beginning of training.

Equation~\ref{eq:td-zero} uses $V^\pi$ as the critic, which is standard with actor-critic methods. This can be done even in offline settings, by using a Temporal Differences loss for the value function \citep{impala}. Unfortunately, this can lead to divergence during offline training, which is a known problem \citep{hasselt2018}. One solution could be early stopping: the policy $\pi$ improves at first before deteriorating, therefore we could stop training early and obtain an improved policy $\pi$. However, this method requires running the environment to detect when to stop, which is contrary to the rules of \scbenchmark{}. Instead, we used $V^\mu$ as a critic, and keep it fixed, instead of $V^\pi$.

\paragraph{Emphatic Offline Actor-Critic (E-OAC) agent.}
\emph{N-step Emphatic Traces (NETD)} \citep{jiang2021} avoids divergence in off-policy learning under some conditions, by weighting the updates beyond the importance sampling ratios. We refer to \cite{jiang2021} for details about the computation of the emphatic traces.



\subsection{MuZero}
\label{sec:muzero}

\mzun{} \citep{muzero_unplugged} adapts \emph{Monte Carlo Tree Search (MCTS)} to the offline setting. It has been successful on a variety of benchmarks \citep{gulcehre2020rl,dulac2019challenges}.
In order to handle the large action space of \starcraft{}, we sample multiple actions from the policy and restrict the search to these actions only, as introduced in \cite{sampled_muzero}. This allows us to scale to the large action space of \starcraft. We used the latent model presented in Section~\ref{sec:architecture}, and similarly to the offline actor-critic agents, we only improved the \texttt{function} and \texttt{delay} from the behavior cloning policy.

\paragraph{MuZero Supervised (MZS) agent.}
Similarly to the Offline Actor-Critic case, training the target policy $\pi$ and estimated value function $V^\pi$ jointly can diverge. In an analog approach, a workaround is to only train the policy to estimate the behavior policy, and use the value and latent model to run MCTS at inference time only. This results in only using the losses for the policy and the value function for MuZero. In other words, the loss is simply the following loss:
\begin{align}
    L^{MuZero}(s) = L^{BC}(s) + L^{MSE}(s)
\end{align}
where the policy and value function are computed using the latent model for all steps except the first one, as shown on Equation~\ref{eq:muzero-model}. Although the loss is similar to standard behavior cloning, using this method can lead to improved performance thanks to the regularization effects of the value function training and the latent model. 

\paragraph{MuZero Supervised with MCTS at inference time (MZS-MCTS) agent.}
The MuZero Unplugged algorithm uses MCTS at training time and inference time. As explained above, policy improvement at training time can lead to divergence. Using MCTS at inference time, on the other hand, is stable and leads to better policies. We use the approach detailed in \cite{sampled_muzero} for the inference.

\section{Experiments}
\label{sec:results}

In this section, we measure the influence of several parameters. For simplicity, we use the win rate against the \veryhard{} bot as the metric for these experiments. Most experiments are run in the behavior cloning  setting.
Due to the cost of running such experiments, we could only train a single model per set of parameters, but the consistency of the conclusions leads us to believe that the results are significant.

Moreover, Section~\ref{sec:reference_agents_evaluation} presents the performance of the reference agents on all \scbenchmark{} metrics, as well as against the original AlphaStar agents from \cite{vinyals2019grandmaster}.

In this section, we call number of learner \emph{steps} the number of updates of the weights on minibatches of rollouts of size $M\times K$. We call number of learner \emph{frames} the total number of observations used by the learner, \emph{i.e.} the number of steps multiplied by $M\times K$.

\subsection{Minibatch and rollout sizes}

The minibatch size $M$ and rollout size $K$ influence the final performance of the models. Table~\ref{tab:bc_batch_size} compares some settings in the case of behavior cloning. In all these experiments, the total number of training frames is $10^{10}$. We found that more data per step --- \emph{i.e.} larger $M\times K$ --- leads to better final performance.

There are unfortunately a few constraints to respect. $M\times K$ cannot be increased indefinitely because of the memory usage. The largest value we could use was $16,384$ or $32,768$, depending on the method. Besides, the Offline Actor-Critic and MuZero methods require $K>1$, and larger values of $K$ stabilize the training.

\begin{table}
    \centering
    \captionof{table}{Behavior cloning performance with different minibatch sizes $M$ and rollout lengths $K$.}
    \vspace{1mm}
    \begin{tabular}{cccc}
        \toprule
        Minibatch size $M$ & Rollout length $K$ & $M\times K$ & win rate vs. \veryhard \\
        \midrule
        8,192 & 1 & 8,192 & 70\% \\
        16,384 & 1 & 16,384 & 79\% \\
        256 & 64 & 16,384 & 79\% \\
        32,768 & 1 & 32,768 & 84\% \\
        \bottomrule
    \end{tabular}
    \label{tab:bc_batch_size}
\end{table}

\subsection{Learning rate}

\begin{figure}
    \centering
    \begin{subfigure}[t]{0.45\textwidth}
      \centering
      \includegraphics[width=\textwidth]{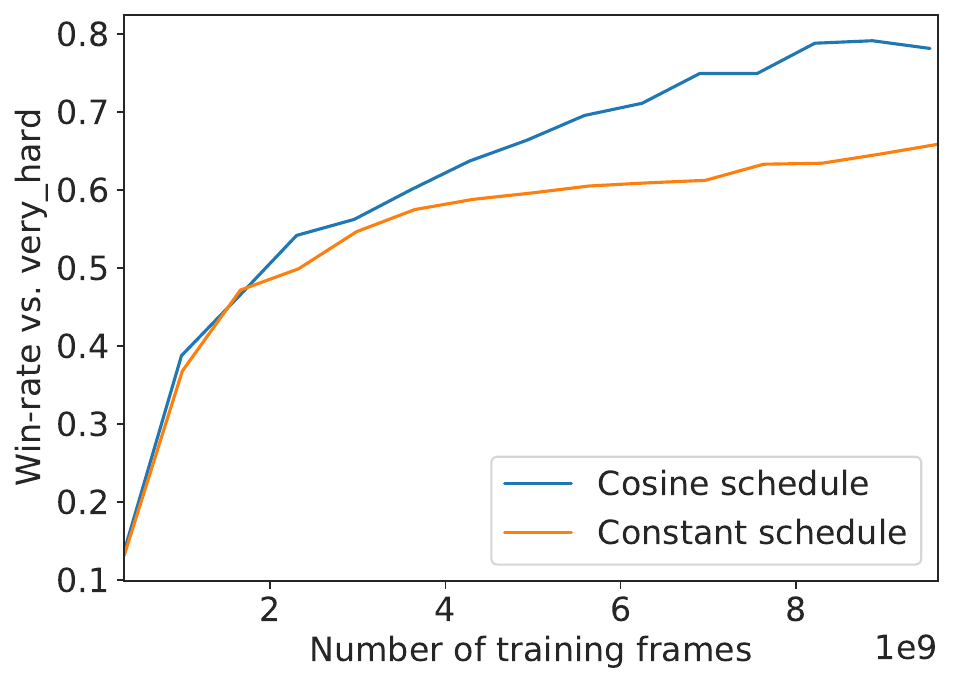}
      \caption{Comparison of learning rate schedules. The constant learning rate, as well as $\lambda_0$, are both set to $5\cdot10^{-4}$.}
      \label{fig:lr_schedule}
    \end{subfigure}
    \hfill
    \begin{subfigure}[t]{0.45\textwidth}
      \centering
      \includegraphics[width=\textwidth]{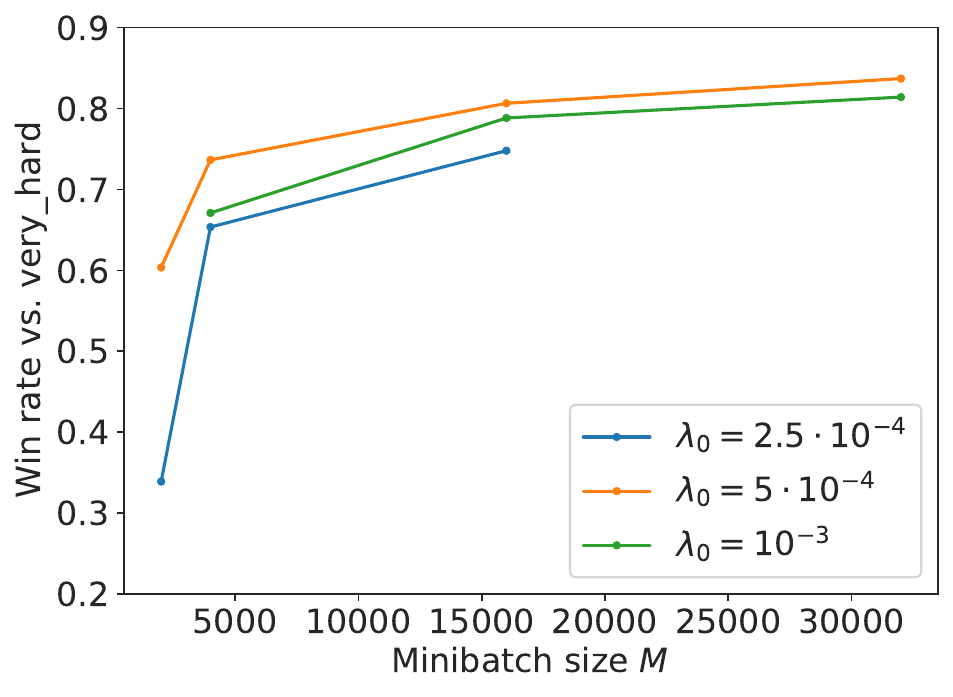}
      \caption{Final performance for different initial learning rates $\lambda_0$ and minibatch sizes $M$.}
      \label{fig:scaling_lr_and_batch_size}
    \end{subfigure}
    \caption{Win rate against the \veryhard{} bot for different learning rate schedules, on behavior cloning.}
\end{figure}

The learning rate $\lambda$ has a significant influence on the final performance of the agents.
We used a cosine learning rate schedule \citep{loshchilov2016sgdr}, parameterized by the initial learning rate $\lambda_0$. Some experiments use a ramp-in period over $N_{\text{ramp-in}}$ frames. At frame $k$, the learning rate is given by
\begin{align}
    \lambda(k) = \min\left(1, \frac{k}{N_{\text{ramp-in}}}\right)\cdot\left(\frac{\lambda_0}{2}\cdot\cos\left(\pi\cdot\frac{k}{k_{max}}\right)+0.5\right)
\end{align}
where $k_{max}$ is the total number of training frames. We compared this schedule to a constant learning rate on Figure~\ref{fig:lr_schedule}.

Figure~\ref{fig:scaling_lr_and_batch_size} shows the final performance for different values for $\lambda_0$ and different minibatch sizes $M$. Since these experiments are slow, it is common to look at the win rate before the experiment is over and decide to compare the results before convergence. It is noteworthy to mention that it should be avoided to find the optimal $\lambda_0$. Indeed, we observed that after only $10^9$ steps, the best performance is obtained with the $\lambda_0=10^{-3}$, but after the full training, it changes.

In the following experiments, we used $\lambda_0=5\cdot 10^{-4}$ unless specified otherwise. The learning rate schedules used to train the reference agents are detailed in Appendix~\ref{app:implementations}.

\subsection{Number of training frames}

\begin{figure}
  \centering
  \begin{subfigure}[t]{0.45\textwidth}
    \centering
    \includegraphics[width=\textwidth]{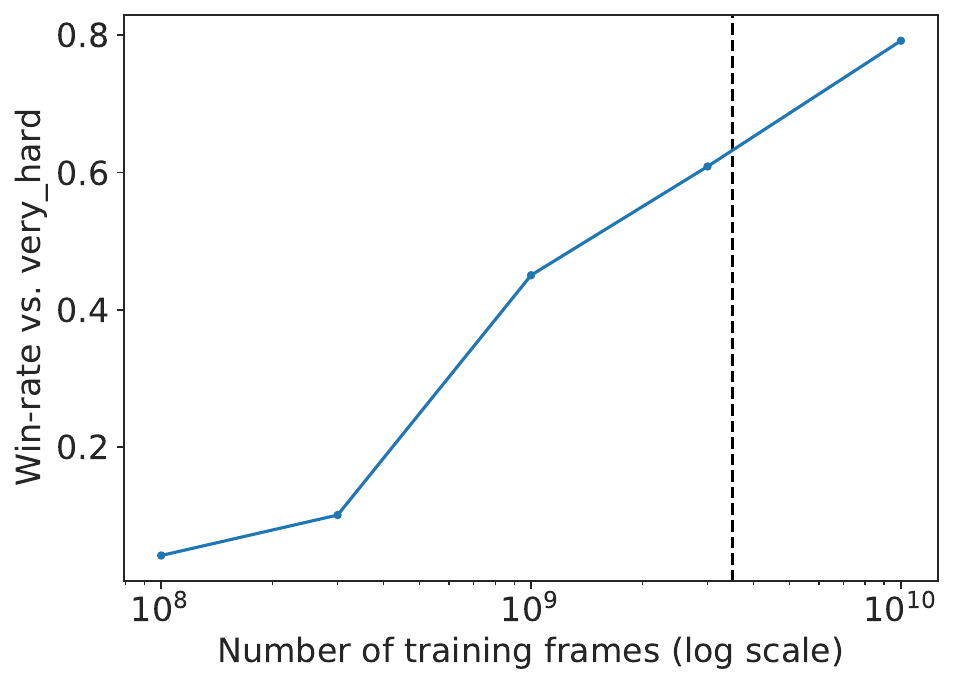}
    \caption{Performance when varying the number of training input frames.}
    \label{fig:scaling_num_observations}
  \end{subfigure}
  \hfill
  \begin{subfigure}[t]{0.45\textwidth}
      \centering
      \includegraphics[width=\textwidth]{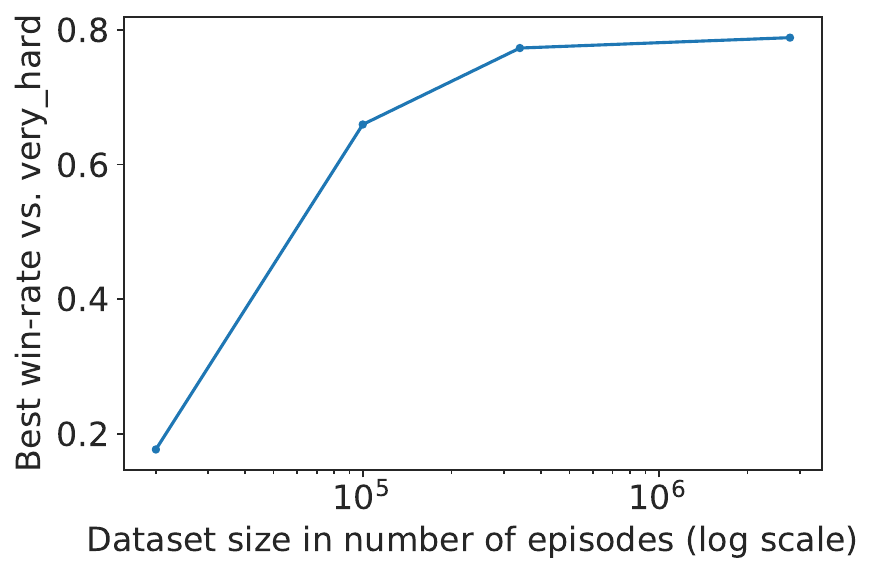}
      \caption{Performance when training on $10^{10}$ frames, with different number of unique episodes in the dataset. Small sizes lead to overfitting so we show the peak win rate over the course of training, instead of the final value.}
      \label{fig:scaling_dataset_size}
  \end{subfigure}
  \caption{Win rate against the \veryhard{} bot when scaling the data.}
\end{figure}

As mentioned in Section~\ref{sec:training}, we trained most of our agents over $k_{max}=10^{10}$ input frames. We measured the behavior cloning performance of the agents trained on fewer frames, as shown on Figure~\ref{fig:scaling_num_observations}. The performance increases logarithmically with the number of training frames.

Note that the fine-tuned (BC-FT) and offline actor-critic (OAC and E-OAC) reference agents were trained on $10^9$ frames, restarting from the behavior cloning agent. Therefore, they were trained on a total on a total of $11$ billion frames, whereas the BC, MZS and MZS-MCTS were only trained on $10$ billion frames.

\subsection{Dataset size}

Figure~\ref{fig:scaling_dataset_size} shows the behavior cloning performance for different dataset sizes, \emph{i.e.} number of unique episodes used for training. For all the points on this curve, we trained the model on the full $k_{max}=10^{10}$ frames, which means that episodes are repeated more often with smaller sized datasets. Unlike most experiments, here we used minibatch size $M=16,384$ and a learning rate of $10^{-3}$.

It is noteworthy that the win rate with only $10\%$ of the episodes in the dataset is close to the best one. This can be used to save storage at very little cost, if it is a concern. However, further reducing the dataset size significantly alters the performance.

\subsection{Data filtering}
\label{sec:dataset_filtering}

\begin{table}
\captionof{table}{Performance of behavior cloning when using different MMR filtering schemes. Higher quality data also means fewer episodes, therefore worse performance. High quality data for fine-tuning gives the best results.}
\vspace{1mm}
    \centering
    \begin{tabular}{ccc|ccc|c}
        \toprule
        \multicolumn{3}{c|}{Main training}&\multicolumn{3}{c|}{Fine-tuning}& \\
        \midrule
        MMR & filter &\#episodes & MMR & filter & \#episodes & win rate vs. \veryhard{} \\
        \midrule
        >3500 & win+loss & 2,776,466 &&&& 84\% \\
        >6000 & win+loss & 64,894 &&&& 65\% \\
        >6000 & win & 32,447 &&&& 51\% \\
        >3500& win+loss &2,776,466 & >6200 & win & 21,836 & \textbf{89\%} \\
        \bottomrule
    \end{tabular}\\
    \label{tab:mmr_comparison}
\end{table}

Filtering the data has a large influence on the final performance of the models. Table~\ref{tab:mmr_comparison} shows that, for behavior cloning, restricting the training set to fewer, higher quality episodes results in poorer performance. However, training using the full dataset followed by a fine-tuning phase on high quality data works best (BC-FT reference agent).

\subsection{Memory}
\label{sec:memory}

The AlphaStar agent of \cite{vinyals2019grandmaster} uses an LSTM module to implement memory. We have tried using LSTM, Transformers and no memory. Surprisingly, we found that no memory performs better than LSTM for behavior cloning, although the final values of the losses are higher.

Results with transformers are more ambivalent. The transformer agent performs similarly to the memory-less agent on $10^{10}$ training frames. However, although the performance of the memory-less agent saturates beyond $k_{max}=10^{10}$ frames, transformers do not, and they outperform the memory-less agent if trained for $2\cdot 10^{10}$ frames. Table~\ref{tab:memory} summarizes the performance of these agents versus the \veryhard{} bot.

\begin{table}
\captionof{table}{Comparison of behavior cloning performance against the \veryhard{} built-in bot with different implementations of memory.}
\vspace{1mm}
    \centering
    \begin{tabular}{c|c}
        \toprule
        Memory & Win rate vs. \veryhard\\
        \midrule
        LSTM & 70\% \\
        No Memory & 84\% \\
        Transformer, $k_{max}=10^{10}$ frames & 85\% \\
        Transformer, $k_{max}=2\cdot{}10^{10}$ frames & 89\% \\
        \bottomrule
    \end{tabular}\\
    \label{tab:memory}
\end{table}

Transformer require extensive hyperparameter tuning and longer training times. Therefore all agents presented in the main experiments are memory-less.
Using transformers for other Offline RL baselines may result in more pronounced benefits and is an interesting future research direction.

%
%


\subsection{Model size}

Because of the complexity of the model, many parts could be scaled individually, but this would be prohibitive. We chose our standard model size as the largest model which can fit in memory without significant slowdown. Scaling down the width of the model by half leads to significant decrease of the performance, from 83\% to 76\% win rate against the \veryhard{} bot, however scaling down the depth by half (rounding up) barely changes the win rate (82\%). In the setup used for our experiments, the training speed does not significantly increase when decreasing the depth, but potential speed gains could be obtained in the future by using smaller models.

\subsection{Temperature and sampling}

During inference, we sample from the policy $\pi(\cdot|s_t)$ given a state $s_t$. In practice, the policy is characterized by a \emph{logits} vector $y$ such that:
\begin{align}
    \pi(a|s_t) = \frac{\exp\left(y_a/\beta\right)}{\sum_{a'=0}^{|\gA|}\exp\left(y_{a'}/\beta\right)}
\end{align}
where $\beta$ is the \emph{temperature}. During training, the temperature is $\beta=1$ but it can be changed at inference time, in order to make the policy more peaked.

We found that $\beta=0.8$ is a good value for the temperature during inference, as shown on Table~\ref{tab:temperature}.

\begin{table}
\captionof{table}{Win rate of different agents versus the \veryhard{} bot with two different sampling temperatures.}
\vspace{1mm}
    \centering
    \begin{tabular}{c|cccc}
        \toprule
        Temperature $\beta$ & \multirow{2}{*}{Behavior Cloning} & \multirow{2}{*}{Fine-Tuning} & \multirow{2}{*}{Offline Actor Critic} & Emphatic \\
        during inference&&&&Offline Actor-Critic\\
        \midrule
        1 & 84\% & 90\% & 93\% & 93\% \\
        0.8 & 88\% & 95\% & 98\% & 97\% \\
        \bottomrule
    \end{tabular}\\
    \label{tab:temperature}
\end{table}

\subsection{Critic of offline actor-critic}

\begin{figure}
    \centering
    \begin{subfigure}[t]{0.45\textwidth}
    \centering
        \includegraphics[width=\textwidth]{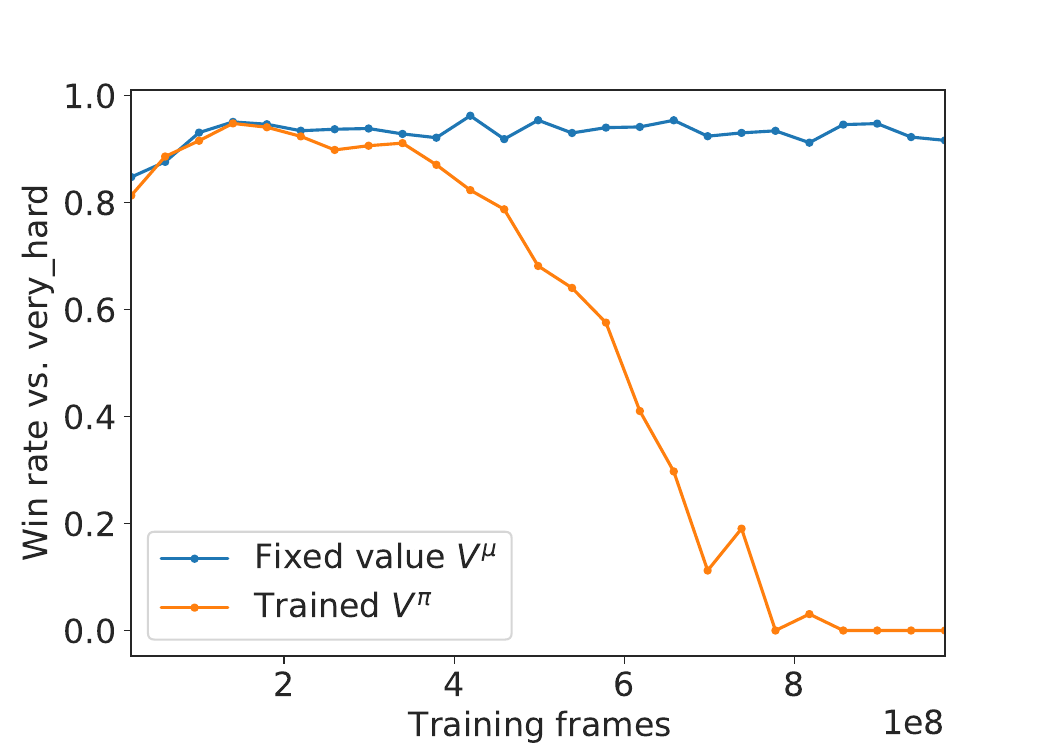}
        \caption{Win rate against the \veryhard{} bot for offline actor-critic training.}
        \label{fig:vpi_vmu_accuracy}
    \end{subfigure}
    \hfill
    \begin{subfigure}[t]{0.45\textwidth}
    \centering
        \includegraphics[width=\textwidth]{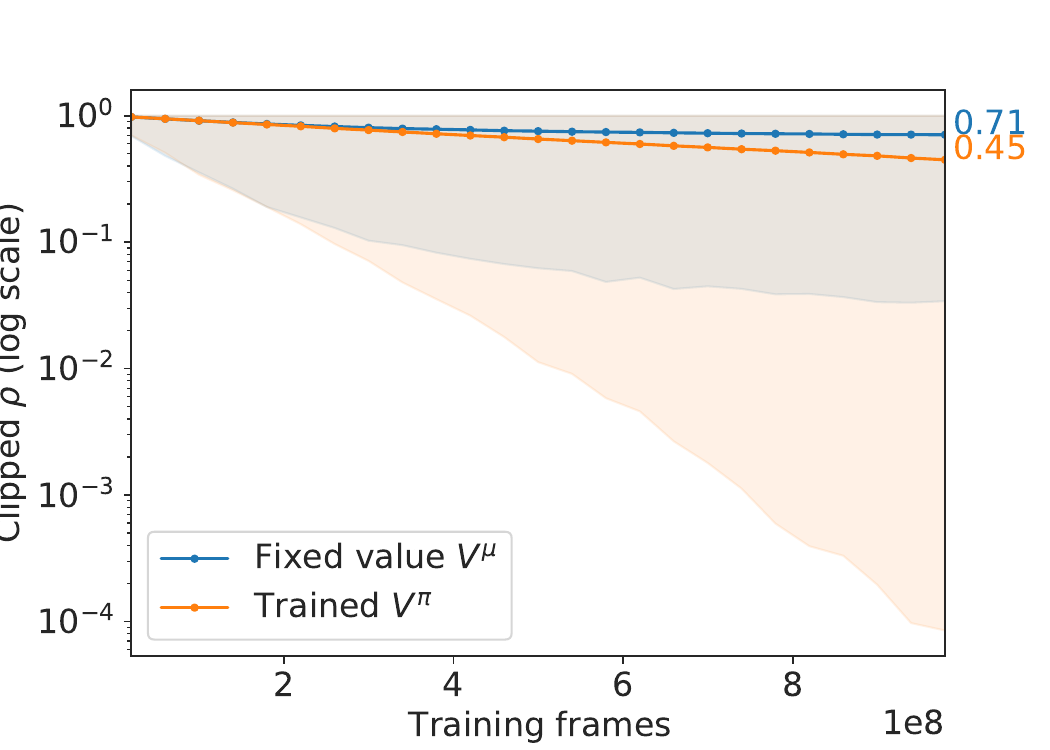}
        \caption{Clipped importance sampling $\rho$ over training.}
        \label{fig:vpi_vmu_rho}
    \end{subfigure}
    \caption{Performance and importance sampling values over offline actor-critic training, comparing $V^\pi$ and $V^\mu$ as the critic.}
    \label{fig:vpi_vmu_comparisons}
\end{figure}

For the offline actor-critic, we experimented with using the value function of the target policy, $V^{\pi}$, as the critic, instead of using the fixed value function of the behavior policy, $V^\mu$. Figure~\ref{fig:vpi_vmu_accuracy} shows the divergence observed when using $V^\pi$. Indeed, although the win rate first increases in both cases, it stays high with $V^\mu$ but deteriorates with $V^\pi$. On Figure~\ref{fig:vpi_vmu_rho}, we can see that the importance sampling $\rho$ (clipped by the V-Trace algorithm) decayed much faster and lower when using $V^\pi$. This means that the policy $\pi$ and $\mu$ got further and further apart on the training set and eventually diverged.

\subsection{MCTS during training and inference}

Our preliminary experiments on using the full MuZero Unplugged algorithm, \emph{i.e.}\ training with MCTS targets, were not successful. We found that the policy would collapse quickly to a few actions with high (over-)estimated value. While MCTS at inference time improves performance, using MCTS at training time leads to a collapsed policy. To investigate this further, we evaluated the performance of repeated applications of MCTS policy improvement on the behavior policy $\hat{\mu}$ and value $V^{\mu}$. We do this by training a new MuZero model using MCTS actions of a behavior policy, i.e. $\hat{\nu} = MCTS(\hat{\mu}, V^{\mu})$. We found that the MCTS performance of this policy $MCTS(\hat{\nu}, V^{\mu})$ is worse than the performance of $\hat{\nu}$ or $MCTS(\hat{\mu}, V^{\mu})$. Thus, repeated applications of MCTS do not continue to improve the policy. We believe this is likely due to MCTS policy distribution generating out of distribution action samples with over-estimated value estimates.

\begin{figure}
    \centering
    \begin{subfigure}[t]{0.45\textwidth}
        \includegraphics[width=\textwidth]{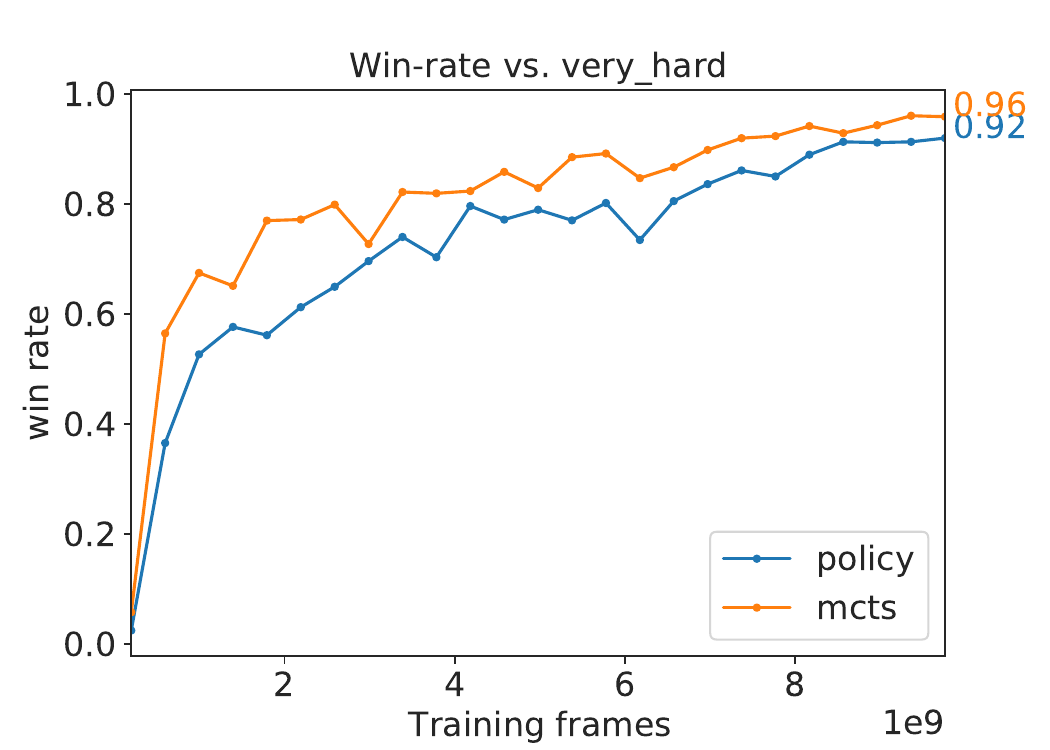}
    \end{subfigure}
    \caption{Comparison of the win rates of the MZS and MZS-MCTS agents over the course of training. Using MCTS outperforms not using it throughout training.}
    \label{fig:mcts_vs_policy}
\end{figure}

Figure~\ref{fig:mcts_vs_policy} compares using MCTS or not during inference. We can see that using MCTS always outperforms not using it, even at the beginning of training.

\subsection{Evaluation of the reference agents}
\label{sec:reference_agents_evaluation}

\definecolor{LightCyan}{rgb}{0.88,1,1}
\definecolor{Gray}{gray}{0.9}
\begin{table}
    \centering
    \captionof{table}{Evaluation of the 6 reference agents with the proposed metrics. Agents highlighted in pale blue utilize offline RL algorithms, whereas the other 3 rely on behavior cloning.  In the bottom portion of this table we show performance of agents from \cite{vinyals2019grandmaster}. Our BC agent is most comparable to AlphaStar Supervised but performs better due to significant tuning improvements. The other AlphaStar agents highlighted in grey have differences which make their performance not directly comparable to ours.}
    \vspace{1mm}
    \begin{tabular}{lccc}
    \toprule
    Agent   & Robustness& Elo& vs \veryhard{}\\
    \midrule
    \rowcolor{LightCyan}
    MuZero Supervised with MCTS at inference time  & 50\% &  1578& 95\%\\
    \rowcolor{LightCyan}
    Emphatic Offline Actor-Critic  & 43\% &  1563& 97\%\\
    \rowcolor{LightCyan}
    Offline Actor-Critic  & 43\%&  1548 & 98\%\\
    Fine-tuned Behavior Cloning  & 36\% &  1485& 95\%\\
    MuZero Supervised & 30\%&  1425 & 92\%\\
    Behavior Cloning  & 25\% &  1380& 88\%\\
    \midrule
    \veryhard{} built-in bot &  3\% &  1000 & 50\% \\
    AlphaStar Supervised & 8\% & 1171 & 75\% \\
    \rowcolor{Gray}
    AlphaStar Supervised (Race specific networks) & 17\% & 1280 & 82\% \\
    \rowcolor{Gray}
    AlphaStar Supervised (Race specific networks + FT)& 44\% & 1545 & 94\% \\
    \rowcolor{Gray}
    AlphaStar Final (Race specific networks + FT + Online Learning) & 100\% & 2968 & 100\% \\
    \bottomrule
    \end{tabular}
    \label{tab:head_to_head}
\end{table}

Table~\ref{tab:head_to_head} shows the performance of our six reference agents using our three metrics: robustness, Elo and win rate versus the \veryhard{} built-in bot. These metrics are explained in Section~\ref{sec:evaluation}.
The three best agents utilize offline RL algorithms (highlighted in pale blue).

\begin{wrapfigure}{r}{0.6\textwidth}
    \centering \includegraphics[width=\linewidth]{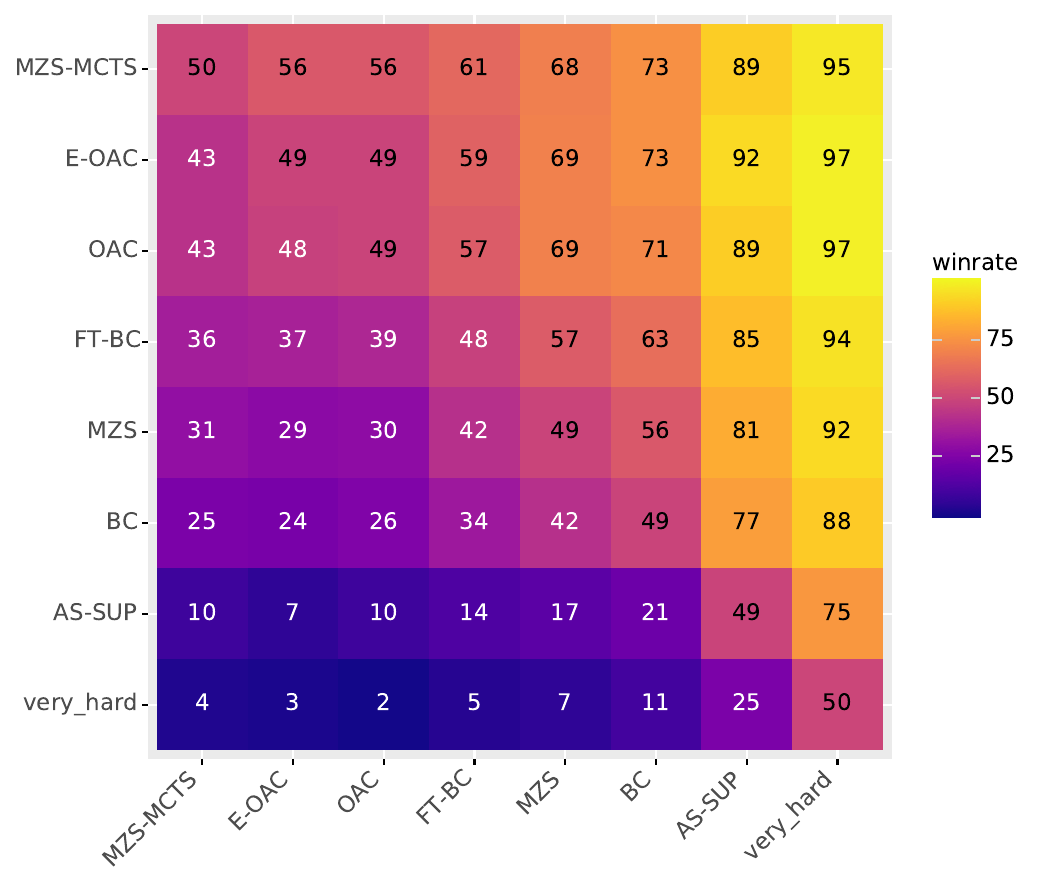}
    \caption{Win rate matrix of the reference agents, normalized between $0$ and $100$. Note that because of draws, the win rates do not always sum to 100 across the diagonal. AS-SUP is the original AlphaStar Supervised agent (not race specific).}
    \label{fig:win_matrix}
\end{wrapfigure}

The full win rate matrix of the reference agents can be seen in Figure~\ref{fig:win_matrix}. A more detailed matrix, split by race, is displayed in Figure~\ref{fig:expanded_win_matrix} in Appendix~\ref{sec:expanded_win_matrix}.

We observe that the MuZero Supervised with MCTS at inference time (MZS) reference agent performs best, although at the cost of slower inference. Generally, we see that the three offline RL methods are ranked closely and significantly higher than behavior cloning. For completeness, we compare with the original AlphaStar agents. The AlphaStar Supervised was trained as three race-specific agents, which is different from the rules of our benchmark (agents should play all races). Therefore, we also compare our agents to a version of AlphaStar Supervised trained to play all races. The win rate of the MZS-MCTS, E-OAC and OAC are 90\%, 93\% and 90\% respectively (see Figure~\ref{fig:win_matrix}). We also note that although offline RL improves upon the behavior cloning baseline, they are far from the online RL performance of AlphaStar Final, which was trained using several orders of magnitude more computing power.

\subsection{Additional offline RL baselines}
\label{sec:additional_results}

We evaluated several typical off-policy and offline RL baselines such as action-value based methods like deep offline Q-Learning \citep{agarwal2020optimistic}, SARSA \citep{rummery1994online}, Critic Regularized Regression (CRR) \citep{wang2020critic}, Batch-Constrained Deep Q-Learning (BCQ) \citep{fujimoto2019off}, Regularized Behavior Value Estimation (R-BVE) \citep{gulcehre2021regularized}, Critic-Weighted Policy (CWP) \citep{wang2020critic} and Return Conditioned Behavior Cloning (RCBC) \citep{srivastava2019training} on \scbenchmark. We also tried Advantage-Weighted Regression (AWR) \citep{peng2019awr}, and Proximal Policy Optimization (PPO) \citep{schulman2017ppo}. None of those approaches could achieve better results than the agents such as BC and FT-BC.
In this section, we will highlight some of those approaches and challenges we faced when we scaled them up to \starcraft.

\paragraph{Deep offline Q-Learning.}
We trained offline Q-learning agents based on DQN \citep{mnih2015human}, that are predicting Q-values and policy with the same output layer for only the \texttt{function} argument. However, the training of those offline Q-learning agents was very unstable, and they have 0\% win rate against the \veryhard{} bot. Moreover, typical approaches to improve Q-learning in the such as N-step returns, dueling network architecture \citep{wang2016dueling} and double-Q-learning \citep{van2016deep} did not improve the performance of our Q-learning agents. Besides the policies themselves, the accuracy of the action-values predicting the returns was poor. 

\paragraph{Offline RL methods using action values.}
We trained CRR, BCQ, CWP, and R-BVE agents with an action-value Q-head on the \texttt{function} argument.
CRR and R-BVE achieved very similar results, and neither could provide significant improvements over the BC agent. BVE and R-BVE were very stable in terms of training. For CRR, we also used BVE to learn the Q-values instead of the. On the other hand, CRR, R-BVE, CWP, and BCQ all achieved around 83-84\% win rate against the \veryhard{} bot. 

\paragraph{Return conditioned behavior cloning. (RCBC)}
We trained a BC agent conditioned on the win-loss return. During inference, we conditioned it on winning returns only, to make it model behavior policy used in winning games. We did not notice any difference, in fact the agent seemed to ignore the return conditioning. We attribute this to the two well-known failure points of RCBC approaches: stochasticity arising due to the noisy opponents, and inability to do trajectory stitching \citep{brandfonbrener2022does}.




\section{Discussion}
\label{sec:discussion}

Behavior cloning is the foundation of all agents in this work. The offline RL agents start by estimating the behavior policy using behavior cloning, then improve upon it using the reward signal. This allows them to perform significantly better than the behavior cloning results. Indeed, although the agents are conditioned on the MMR during training, the behavior cloning agents are still fundamentally limited to estimating the behavior policy, ignorant about rewards. As a result, the policy they learn is a smoothed version of all the policies that generated the dataset. In contrast, offline RL methods use rewards to improve learned policies in different ways. Offline Actor-Critic methods use policy-gradient. MCTS at inference time aims at maximizing the estimated return. Even the MuZero Supervised without MCTS and the fine-tuned behavior cloning make use of the reward, and outperform the BC baseline.

We have observed that algorithms originally designed for online learning --- even with off-policy corrections --- do not work well when applied directly to the full offline RL setting. We attribute this in part to the problem of the \emph{deadly triad} \citep{Tsitsiklis97, sutton2018, hasselt2018}. However, many recent works have found these algorithms can be made more effective simply by making modifications that ensure the target policy stays close to the behavior policy $\mu$, that the value function stays close to $V^{\mu}$, or both. Our results with Actor-Critic and MuZero are in accordance with these findings.

Among all the methods we tried, the reference agents are the ones which led to improved performance. However, we have tried several other methods without success, listed in Section~\ref{sec:additional_results}. We may have failed to find the modifications which would have made these methods perform well on this dataset. However, \scbenchmark{} is fundamentally difficult:

\begin{itemize}
    \item \textbf{Limited coverage.} The action space is very large, the state-action coverage in the dataset is low, and the environment is highly partially observable. This makes it challenging for the value function to extrapolate to unseen state and actions \citep{fujimoto2019off,gulcehre2022empirical}. This is particularly impactful for Q-values based methods, since there are significantly fewer states than state-action pairs, so it is easier to learn state value functions. 
    The approaches like R-BVE mitigate extrapolation errors during training, but still the agent has to extrapolate during inference. 
    \item \textbf{Weak learning signal.} The win-loss reward is a weak learning signal to learn a good policy because the dataset has games with a wide range of qualities and the win-loss reward ignores this. For example, the winner of a game between two low-skilled players would consistently lose to the loser of a game between two professional players. Thus, purely relying on the win-loss signal in the offline RL case is problematic.
    \item \textbf{Credit assignment} is difficult due to large action space, sparse rewards, long-horizon, and partial observability. This exacerbates the problems with the offline RL based agents.
    \item \textbf{Autoregressive action space} requires learning autoregressive Q-values which is challenging and understudied in the literature. In this paper, we side-stepped this by just learning a Q-function only for the \texttt{function} argument.
\end{itemize}

\section{Related work}
\label{sec:related}

Online RL has been very impactful for building agents to play computer games. RL agents can outperform professional human players in many games such as \starcraft{} \citep{vinyals2019grandmaster}, DOTA \citep{berner2019dota} or Atari \citep{mnih2015human,badia2020agent57}. Similar levels of progression have been observed on board games, including chess and Go \citep{silver2016mastering,silver2017mastering}. Although offline RL approaches have shown promising results on Atari recently \citep{schrittwieser2021online}, they have not been previously applied on complex partially observable games using data derived from human experts.

RL Unplugged \citep{gulcehre2020rl} introduces a suite of benchmarks for Offline RL with a diverse set of task domains with a unified API and evaluation protocol. D4RL \citep{fu2020d4rl} is an offline RL benchmark suite focusing only on mixed data sources. However, both RL Unplugged and D4RL lack high-dimensional, partially observable tasks. This paper fills that gap by introducing a benchmark for \starcraft{}.


Offline RL has become an active research area, as it enables us to leverage fixed datasets to learn policies to deploy in the real-world. Offline RL methods include 1) policy-constraint approaches that regularize the learned policy to stay close to the behavior policy \citep{wang2020critic,fujimoto2019off}, 2) value-based approaches that encourage more conservative value estimates, either through a pessimistic regularization or uncertainty \citep{kumar2020conservative,gulcehre2021regularized}, 
3) model-based approaches \citep{yu2020mopo,kidambi2020morel,schrittwieser2021online}, and 4) adaptations of standard off-policy RL methods such as DQN \citep{agarwal2020optimistic} or D4PG \citep{wang2020critic}. Recently methods using only one-step of policy improvement has been proven to be very effective on offline reinforcement learning \citep{gulcehre2021regularized,brandfonbrener2021offline}.




\section{Conclusions}
\label{sec:conclusions}

Offline RL has enabled the deployment of RL ideas to the real world. Academic interest in this area has grown and several benchmarks have been proposed, including RL-Unplugged \citep{gulcehre2020rl}, D4RL \citep{fu2020d4rl}, and RWRL \citep{dulac2019challenges}. However, because of the relatively small-scale and synthetic nature of these benchmarks, they don't capture the challenges of real-world offline RL.

In this paper, we introduced \scbenchmark{}, a benchmark to evaluate agents which play \starcraft{} by learning only from \emph{offline} data. This data is comprised of over a million games games mostly played by amateur human \starcraft{} players on Blizzard's Battle.Net.\footnote{\url{https://en.wikipedia.org/wiki/Battle.net}} Thus, the benchmark more accurately captures the challenges of offline RL where an agent must learn from logged data, generated by a diverse group of weak experts, and where the data doesn't exhaust the full state and action space of the environment.


We showed that offline RL algorithms can exceed 90\% win rate against the all-races version of the previously published AlphaStar Supervised agent (trained using behavior cloning). However, the gap between online and offline methods still exists and we hope the benchmark will serve as a testbed to advance the state of art in offline RL algorithms. 

%
%
\subsubsection*{Acknowledgments}
We would like to thank Alistair Muldal for helping with several aspects of the open-sourcing which went a long way in making the repository user-friendly. We would like to thank Scott Reed and David Silver for reviewing the manuscript, the AlphaStar team \citep{vinyals2019grandmaster} for sharing their knowledge and experience about the game. We would like to thank the authors of \mzun{} \citep{muzero_unplugged} and \mzsa{} \citep{sampled_muzero} for advising on the development of the \mzsup{} agent. We also thank the wider DeepMind research, engineering, and environment teams for the technical and intellectual infrastructure upon which this work is built. We are grateful to the developers of tools and frameworks such as JAX \citep{deepmind2020jax}, Haiku \citep{haiku2020github} and Acme \citep{hoffman2020acme} that enabled this research.

\bibliography{refs}
\bibliographystyle{tmlr}
\vfill

\pagebreak{}

\input{appendix.tex}

\end{document}

%% file: math_commands.tex
\usepackage{amsmath,amsfonts,bm}

\newcommand*{\defeq}{\stackrel{\text{def}}{=}}

\def\eqref#1{equation~\ref{#1}}

\def\1{\bm{1}}

\DeclareMathAlphabet{\mathsfit}{\encodingdefault}{\sfdefault}{m}{sl}
\SetMathAlphabet{\mathsfit}{bold}{\encodingdefault}{\sfdefault}{bx}{n}

\def\gA{{\mathcal{A}}}

\def\gI{{\mathcal{I}}}

\def\gL{{\mathcal{L}}}

\def\gS{{\mathcal{S}}}

\def\sR{{\mathbb{R}}}



%% file: defns.tex
\usepackage{mathtools}
\usepackage{dsfont}
\usepackage{booktabs}
\usepackage{xfrac}
\usepackage{bbm}
\usepackage{booktabs}
\usepackage{algorithm}
\usepackage{algorithmic}

\usepackage[most]{tcolorbox}
\usepackage{xparse}
\usepackage{lipsum}
\usepackage{changepage}
\usepackage{enumitem}
\usepackage{graphicx}
\usepackage{xcolor}
\usepackage{wrapfig}
\usepackage{amsmath,amssymb,amsfonts}

\usepackage{caption}

\newif\ifcomments
\commentsfalse
\commentstrue  
\ifcomments
\newcommand{\cg}[1]{\textcolor{purple}{\bf\small [#1 --CG]}}

\newcommand{\mati}[1]{\textcolor{green}{\bf\small [#1 --MM]}}
\definecolor{darkblue}{RGB}{0, 0, 83}
\newcommand{\kmh}[1]{\textcolor{darkblue}{\bf\small [#1 --KMH]}}
\newcommand{\razp}[1]{\textcolor{magenta}{\bf\small [#1 --razvan]}}
\newcommand{\razpt}[1]{\textcolor{magenta}{\bf\small #1}}
\newcommand{\md}[1]{\textcolor{pink}{\bf\small [#1 -- MD]}}
\newcommand{\ar}[1]{\textcolor{orange}{\bf\small [#1 -- ali]}}
\newcommand{\yc}[1]{\textcolor{blue}{\bf\small [#1 --YC]}}

\else
\newcommand{\cg}[1]{}
\newcommand{\mati}[1]{}
\newcommand{\kmh}[1]{}
\newcommand{\razp}[1]{}
\newcommand{\razpt}[1]{}
\newcommand{\md}[1]{}
\newcommand{\ar}[1]{}
\newcommand{\yc}[1]{}
\newcommand{\todo}[1]{}
\fi

\newcommand{\squishlist}{
   \begin{list}{$\bullet$}
    { \setlength{\itemsep}{0pt}      \setlength{\parsep}{3pt}
      \setlength{\topsep}{3pt}       \setlength{\partopsep}{0pt}
      \setlength{\leftmargin}{1.5em} \setlength{\labelwidth}{1em}
      \setlength{\labelsep}{0.5em} } }

\newcommand{\squishlisttwo}{
   \begin{list}{$\bullet$}
    { \setlength{\itemsep}{0pt}    \setlength{\parsep}{0pt}
      \setlength{\topsep}{0pt}     \setlength{\partopsep}{0pt}
      \setlength{\leftmargin}{2em} \setlength{\labelwidth}{1.5em}
      \setlength{\labelsep}{0.5em} } }

\newcommand{\squishend}{
    \end{list}  }

\newcommand{\two}{(2)}

\newcommand{\calD}{{\cal D}}

\newcommand{\data}{\calD}

\newcommand{\trans}{P}

\DeclareMathAlphabet{\mathpzc}{OT1}{pzc}{m}{n}

%% file: appendix.tex
\appendix

\section{Appendix}

\subsection{StarCraft II Interface}
\label{app:sc2_interface}

\starcraft{} features large maps on which players move their units. They can also construct buildings (units which cannot move), and gather resources. At any given time, they can only observe a subset of the whole map through the \emph{camera}, as well as a coarse, zoomed-out version of the whole map called the \emph{minimap}. In addition, units have a \emph{vision} field such that any unit owned by the opponent is hidden unless it is in a vision field. Human players play through the \emph{standard interface}, and receive some additional information, such as the quantity of resources owned or some details about the units currently selected, and audible cues about events in the game. Players issue orders by first selecting units, then choosing an ability, and lastly, for some actions, a target, which can be on the camera or the minimap.

While human players use their mouse and keyboard to play the game, the agents use an API, called the \emph{raw interface}, which differs from the standard interface.\hspace{-0.6em}\footnote{Although the raw interface is designed to be as fair as possible when compared to the standard interface, in particular the observation and actions do not contain significantly different information between the two.} In this interface, observations are split into three modalities, namely \texttt{world}, \texttt{units} and \texttt{vectors} which are enough to describe the full observation:
\begin{itemize}
    \item \textbf{World.} A single tensor called \emph{world} which corresponds to the minimap shown to human players. It is observed as 8 feature maps with resolution 128x128. The feature maps are:
    \begin{itemize}
        \item \texttt{height\_map}: The topography of the map, which stays unchanged throughout the game.
        \item \texttt{visibility\_map}: The area within the vision of any of the agent's units.
        \item \texttt{creep}: The area of the map covered by Zerg "creep".
        \item \texttt{player\_relative}: For each pixel, if a unit is present on this pixel, this indicates whether the unit is owned by the player, the opponent, or is a neutral construct.
        \item \texttt{alerts}: This encodes alerts shown on the minimap of the standard interface, for instance when units take damage.
        \item \texttt{pathable}: The areas of the map which can be used by ground units.
        \item \texttt{buildable}: The areas of the map which where buildings can be placed.
        \item \texttt{virtual\_camera}: The area currently covered by the virtual camera. The virtual camera restricts detailed vision of many \texttt{unit} properties, as well as restricts some actions from targeting units/points outside the camera. It can be moved as an action.
    \end{itemize}
    \item \textbf{Units.} A list of units observed by the agent. It contains all of the agent's units as well as the opponent's units when within the agent's vision and the last known state of opponent's buildings. For each unit, the observation is a vector of size 43 containing all the information that would be available in the game standard interface. In particular, some of the opponent's unit information are masked if they are not in the agent's virtual camera. In addition, the list also contains entries for \emph{effects} which are temporary, localized events in the game (although effects are not units in a strict sense, they can be represented as such). In our implementation, this list can contain up to 512 entities. In the rare event that more than 512 units/effects exist at once, the additional observations will be truncated and not visible to the agent.
    \item \textbf{Vectors.} Global inputs are gathered into \texttt{vectors}. They are:
    \begin{itemize}
        \item \texttt{player\_id}: The id of the player (0 or 1). This is not useful to the agent.
        \item \texttt{minerals}: The amount of minerals currently owned.
        \item \texttt{vespene}: The amount of vespene gas currently owned.
        \item \texttt{food\_used}: The current amount of food currently used by the agent's units. Different units use different amount of food and players need to build structures to raise the \texttt{food\_cap}.
        \item \texttt{food\_cap}: The current amount of food currently available to the agent.
        \item \texttt{food\_used\_by\_workers}: The current amount of food currently used by the agent's workers. Workers are basic units which harvest resources and build structures, but rarely fight.
        \item \texttt{food\_used\_by\_army}: The current amount of food currently used by the agent's non-worker units.
        \item \texttt{idle\_worker\_count}: The number of workers which are idle. Players typically want to keep this number low.
        \item \texttt{army\_count}: The number of units owned by the agent which are not workers.
        \item \texttt{warp\_gate\_count}: The number of warp gates owned by the agent (if the agent's race is Protoss).
        \item \texttt{larva\_count}: The number of larva currently available to the agent (if the agent's race is Zerg).
        \item \texttt{game\_loop}: The number of internal game steps since the beginning of the game.
        \item \texttt{upgrades}: The list of upgrades currently unlocked by the agent.
        \item \texttt{unit\_counts}: The number of each unit currently owned by the agent. This information is contained in the units input, but represented in a different way here.
        \item \texttt{home\_race}: The race of the agent.
        \item \texttt{away\_race}: The race of the opponent. If the opponent is has chosen a random race, it is hidden until one of their unit is observed for the first time.
        \item \texttt{prev\_delay}: The number of internal game steps since the last observation. During inference, this can be different from the \texttt{delay} argument of the previous action (see Section~\ref{section:dataset}), for instance if there was lag.
    \end{itemize}
\end{itemize}

Each action from the raw interface combines up to three standard actions: unit selection, ability selection and target selection. In practice, each raw action is subdivided into up to 7 parts, called \emph{arguments}. It is important to note that the arguments are not independent of each other, in particular the \texttt{function} determines which other arguments are used. The arguments are detailed below:
\begin{itemize}
    \item \textbf{Function.} This corresponds to the ability part of the \starcraft{} API, and specifies the action. Examples include: \texttt{Repair}, \texttt{Train\_SCV}, \texttt{Build\_CommandCenter} or \texttt{Move\_Camera}.
    \item \textbf{Delay.} In theory, the agent could take an action at each environment step. However, since \starcraft{} is a real-time game, the internal game steps are very quick\footnote{22.4 steps per second.} and therefore it would not be fair to humans, which cannot issue action that fast. In addition, it would make episodes extremely long which is challenging to learn. Therefore the agent specifies how many environment steps will occur before the next observation-action pair. Throttling is used to make sure the agent cannot issue too many actions per second.
    \item \textbf{Queued.} This argument specifies whether this action should be applied immediately, or queued. This corresponds to pressing the Shift key in the standard game interface.
    \item \textbf{Repeat.} Some repeated identical actions can be issued very quickly in the standard interface, by pressing keyboard keys very fast, sometimes even issuing more than one action per internal game step. The repeat argument lets the agent repeat some actions up to 4 times in the same step. This is mainly useful for building lots of Zerg units quickly.
    \item \textbf{Unit tags.} This is the equivalent of a selection action in the standard interface. This argument is a mask over the agent's units which determines which units are performing the action. For instance for a \texttt{Repair} action, the unit tags argument specify which units are going to perform the repair.
    \item \textbf{Target unit tag.} Which unit an action should target. For instance, a \texttt{Repair} action needs a specific unit/building to repair. Some actions (e.g. \texttt{Move}) can target either a unit or a point. Those actions are split into two functions. There are no actions in \starcraft{} that target more than one unit (some actions can affect more than one unit, but those actions specify a target point).
    \item \textbf{World.} Which point in the world this action should target. It is a pair of $(x, y)$ coordinates aligned with the world observation. For example \texttt{Move\_Camera} needs to know where to move the camera.
\end{itemize}

\subsection{Evaluation Metrics}
\label{app:eval}
Let us assume that we are given an agent to evaluate $\mathrm{p}$ and a collection of reference agents $$\mathrm{Q} = \{\mathrm{q}_j\}_{j=1}^N.$$ Each of these players can play all three races of StarCraft II: $$\mathrm{R} = \{\mathrm{terran, protoss , zerg}\}.$$ We define an outcome of a game between a player $\mathrm{p}$ and a reference player $\mathrm{q}$ as 
$$
\mathrm{f}(\mathrm{p}, \mathrm{q}) \defeq \mathbb{E}_{\mathrm{r_p,r_q} \sim U(\mathrm{R})} \mathrm{P}[\mathrm{p\; wins\; against\; q} | \mathrm{r(p)=r_p, r(q)=r_q}],
$$
where $\mathrm{r}(\cdot)$ returns a race assigned to a given player, and the probability of winning is estimated by playing matches over uniformly samples maps and starting locations.

\subsubsection{Robustness computation}
We define robustness of an agent $\mathrm{p}$ with respect to reference agents $\mathrm{Q}$ as 
$$
\mathrm{robustness}_\mathrm{Q}(\mathrm{p}) \defeq 1- \min_{\mathrm{q} \in \mathrm{Q}} \mathrm{f}(\mathrm{p}, \mathrm{q}).
$$
Note, this is 1 minus exploitability~\cite{davis2014using}, simply flipped so that we maximise the score. In particular, Nash equlibrium would maximise this metric, if $\mathrm{Q}$ contained every mixed strategy in the game.

\subsubsection{Elo computation}
We follow a standard Chess Elo model~\cite{elo1978rating}, that tries to predict $\mathrm{f}$ by associating each of the agents with a single scalar (called Elo rating) $\mathrm{e}(\cdot) \in \mathbb{R}$ and then modeling outcome with a logistic model:
$$
\widehat{\mathrm{f}_\mathrm{Elo}}(\mathrm{p}, \mathrm{q}) \defeq \frac{1}{1 + 10^{[\mathrm{e}(\mathrm{p}) - \mathrm{e}(\mathrm{q})]/400}}.
$$
For consistency of the evaluation we have precomputed ratings $\mathrm{e(q)}$ for each $\mathrm{q} \in \mathrm{Q}$. Since ratings are invariant to translation, we anchor them by assigning $\mathrm{e}(\mathrm{very\_hard}) := 1000$.

In order to compute rating of a newly evaluated agent $\mathrm{p}$ we minimise the cross entropy between true, observed outcomes $\mathrm{f}$ and predicted ones (without affecting the Elo ratings of reference agents):
$$
\mathrm{Elo}_\mathrm{Q}(\mathrm{p}) \defeq \arg\min_{\mathrm{e}(\mathrm{p})} \left [ - \sum_\mathrm{q} \mathrm{f}(\mathrm{p}, \mathrm{q}) \log \left ( \widehat{\mathrm{f}_\mathrm{Elo}}(\mathrm{p}, \mathrm{q}) \right ) \right ]=  \arg\max_{\mathrm{e}(\mathrm{p})} \sum_\mathrm{q} \mathrm{f}(\mathrm{p}, \mathrm{q}) \log \left ( \widehat{\mathrm{f}_\mathrm{Elo}}(\mathrm{p}, \mathrm{q}) \right ).
$$
Note, that as it is a logistic model, it will be ill defined if $\mathrm{f}(\mathrm{p}, \mathrm{q}) = 1$ (or 0) for all $\mathrm{q}$ (one could say Elo is infinite). In such situation the metric will be saturated, and one will need to expand $\mathrm{Q}$ to continue research progress. 




\subsection{Training of reference agents}
\label{app:implementations}

In this section, we present the details of the training of the reference agents. We list the hyperparameters and propose pseudocode implementations for each agent.

In the pseudo-code, the data at a time step $t$ is denoted as $X$. We refer to different part of the data $X$: $X_s$ is the observation at time $t$, $X_a$ is the action at time $t$, $X_r$ is the reward at time $t$, $X_R$ is the MC return (in the case of \starcraft{}, this is equal to the reward on the last step of the episode), and $X_{\text{game loop delta}}$ is the number of internal game steps between $t$ and $t+1$ (and 0 on the first step).

\subsubsection{Behavior Cloning (BC)}
\label{app:bc}

This agent is trained to minimize the behavior cloning loss $L^{BC}$ (see Section~\ref{sec:bc}), with weight decay. We used a cosine learning rate schedule with $\lambda_0=5\cdot 10^{-4}$ and no ramp-in, over a total of $k_{max}=10^{10}$ training frames. We used a rollout length $K=1$ and a minibatch size $M=32,768$. We used the Adam optimizer \citep{loshchilov2019decoupled}, and we clip the gradients to $10$ before applying Adam.

\begin{algorithm}
\caption{Behavior Cloning (with rollout length $K$ set to $1$)\label{alg:bc}}
\begin{algorithmic} 
\STATE \textbf{Inputs:} A dataset of trajectories $\mathcal{D}$, a mini batch size $M$, an initial learning rate $\lambda$, the total number of observations processed $n_\mathrm{frames}$, and the initial weights used $\theta$ to parameterise the estimated policy $\hat\mu_\theta$.
\FOR {$i=0..n_\mathrm{frames}/M-1$}
  \STATE{Set the gradient accumulator $g_\mathrm{acc} \leftarrow 0$.}
  \FOR {$j=0..M-1$}
    \STATE {Sample a trajectory $T\sim\mathcal{D}$}
    \STATE {Sample $k$ in $[0, length(T)-K]$}
    \STATE {Set $X \leftarrow T[k]$}
    \STATE{Set $g_\mathrm{acc} \leftarrow g_\mathrm{acc} + \frac{1}{M} \cdot \left(\frac{\partial L_\mathrm{cross\ entropy} (\hat\mu_{\theta}(\cdot|X_s), X_a)}{\partial \theta} \right)$,\\ where $X_s$, $X_a$ are the observation and action parts of $X$, respectively.}
  \ENDFOR
  \STATE{Set $\lambda_i \leftarrow \frac{\lambda}{2}\left(\cos\left(\frac{i\pi}{n_\mathrm{frames}/M-1}\right) + 1 \right)$}
  \STATE{Set $\theta \leftarrow \theta - \lambda_i \cdot Adam(g_\mathrm{acc})$}
\ENDFOR
\RETURN{$\theta$}
\end{algorithmic}
\end{algorithm}

\subsubsection{Fine-Tuned Behavior Cloning (FT-BC)}

For the FT-BC agent, we initialized the weights using the trained BC agent. We then trained it to minimize $L^{BC}$ on $k_{max}=10^9$ frames, with a cosine learning rate schedule with $\lambda_0=10^{-5}$ and $N_{\text{ramp-in}}=10^8$. We used a rollout length $K=1$ and a minibatch size $M=32,768$. The data was restricted to episodes with $MMR>6200$ and reward $r=1$. We used the Adam optimizer, and we clip the gradients to $10$ before applying Adam.

\subsubsection{Offline Actor-Critic (OAC)}
\label{app:oac}

For the OAC agent, we first trained a behavior cloning agent with a value function by minimizing
\begin{align}
    10\cdot{}L^{MSE} + L^{BC}
\end{align}
with the same parameters are the BC agent training, except that weight decay was disabled.

From this model, we then trained using the offline actor-critic loss, minimizing $L^{VTrace}$ for $K_{max}=10^{9}$ frames, using a rollout length $K=64$ and a minibatch size $M=512$. We found that it is important to clip the gradients to $10$ after applying Adam during this phase. We used a per-internal game step discount $\gamma=0.99995$.

The loss $L^{VTrace}$ is the policy loss defined in \cite{impala}, and a a bit more complex than $L^{TD(0)}$ presented in Section~\ref{sec:oac}. We used mixed n-steps TD, with n between 32 and 64. As part of the V-Trace computation, we clipped $\rho$ and $c$ to $1$ (see \cite{impala}).

Value function reaches 72\% accuracy, which is computed as the fraction of steps where the sign of $V^\mu$ is the same as $R$.

Divergence: We observe that when doing so, using the value function $V^\pi$ leads to divergence during training, as shown on Figure~\ref{fig:vpi_vmu_comparisons} in the Appendix.


\begin{algorithm}
\caption{Behavior Cloning with value function training (with rollout length $K$ set to $1$)\label{alg:bc_value}}
\begin{algorithmic} 
\STATE \textbf{Inputs:} A dataset of trajectories $\mathcal{D}$, a mini batch size $M$, an initial learning rate $\lambda$, the total number of observations processed $n_\mathrm{frames}$, and the initial weights used $\theta$ to parameterise the estimated policy $\hat\mu_\theta$ and value function $V^{\hat\mu_\theta}$.
\FOR {$i=0..n_\mathrm{frames}/M-1$}
  \STATE{Set the gradient accumulator $g_\mathrm{acc} \leftarrow 0$.}
  \FOR {$j=0..M-1$}
    \STATE {Sample a trajectory $T\sim\mathcal{D}$}
    \STATE {Sample $k$ in $[0, length(T)-K]$}
    \STATE {Set $X \leftarrow T[k]$}
    \STATE{Set $g_\mathrm{acc} \leftarrow g_\mathrm{acc} + \frac{1}{M} \cdot \left(\frac{\partial L_\mathrm{cross\ entropy} (\hat\mu_{\theta}(\cdot|X_s), X_a)}{\partial \theta} + \frac{\partial L_\mathrm{MSE}(V^{\hat\mu}(X_s), X_r)}{\partial \theta}\right)$,\\ where $X_s$, $X_a$ are the observation and action parts of $X$, respectively.}
  \ENDFOR
  \STATE{Set $\lambda_i \leftarrow \frac{\lambda}{2}\left(\cos\left(\frac{i\pi}{n_\mathrm{frames}/M-1}\right) + 1 \right)$}
  \STATE{Set $\theta \leftarrow \theta - \lambda_i \cdot Adam(g_\mathrm{acc})$.}
\ENDFOR
\RETURN{$\theta$}
\end{algorithmic}
\end{algorithm}

\begin{algorithm}
\caption{Offline Actor-Critic (with fixed critic $V^\mu$)\label{alg:oac}}
\begin{algorithmic}
\STATE \textbf{Inputs:} A dataset of trajectories $\mathcal{D}$, the mini batch size $M$, the rollout length $K$, an initial learning rate $\lambda$, the weights of the estimated behavior policy and value function from BC $\theta_0$ (such that $\hat\mu_{\theta_0}$ is the BC policy, and $V^{\hat\mu_{\theta_0}}$ is the behavior value function), the total number of observations processed $n_\mathrm{frames}$, the bootstrap length $N$, the IS threshold $\bar{\rho}$, a per-game step discount $\gamma_0$.
\STATE {Set $\theta \leftarrow \theta_0$}
\FOR {$i=0..n_\mathrm{frames}/M-1$}
  \STATE{Set the gradient accumulator $g_\mathrm{acc} \leftarrow 0$.}
  \FOR {$j=0..M-1$}
    \STATE {Sample a trajectory $T\sim\mathcal{D}$}
    \STATE {Sample $k$ in $[0, length(T)-K]$}
    \STATE {Set $X \leftarrow T[k:k+K-1]$}
    \STATE{\emph{Compute the TD errors $\delta$ and clipped IS ratios $\bar\rho$ with clipping threshold $\hat\rho$}}
    \FOR {$t=0..K-2$}
      \STATE{Set $\delta[t] \leftarrow X_r[t+1] + \gamma_{t+1}V^{\hat\mu_{\theta_0}}(X_s[t+1])-V^{\hat\mu_{\theta_0}}(X_s[t])$}
      \STATE{Set $\bar{\rho}[t] \leftarrow \min(\bar{\rho}, \frac{\pi_\theta(X_a[t]|X_s[t])}{\hat\mu_{\theta_0}(X_a[t]|X_s[t])})$}
      \STATE{Set $\gamma[t] \leftarrow \gamma_0^p$ where $p=X_\text{game\_loop\_delta}[t]$.}
    \ENDFOR
    \STATE{\emph{Compute the V-Trace targets $v$:}}
    \FOR {$t=0..K-N-1$}
      \STATE{Set $v[t+1] \leftarrow V^{\hat\mu_{\theta_0}}(X_s[t+1]) + \sum_{u=t}^{t+N-1} (\prod_{k=t}^{u-1} \bar{\rho}[k] \gamma[k+1] ) \  \bar{\rho}[u] \delta[u]$}
    \ENDFOR
    \STATE{Set $g_\mathrm{acc} \leftarrow g_\mathrm{acc} + \sum_{t=0}^{N-1}\bar{\rho}[t] (X_r[t+1] + \gamma[t+1] v[t+1]  - V^{\hat\mu_{\theta_0}}(X_s[t]) \frac{\partial\log \pi_{\theta}(X_a[t]| X_s[t])}{\partial \theta}$.}
  \ENDFOR
  \STATE{Set $\lambda_i \leftarrow \frac{\lambda}{2}\left(\cos\left(\frac{i\pi}{n_\mathrm{frames}/M-1}\right) + 1 \right)$}
  \STATE{Set $\theta \leftarrow \theta - \lambda_i \cdot Adam(g_\mathrm{acc})$}
\ENDFOR
\RETURN{$\theta$}
\end{algorithmic}
\end{algorithm}

\subsubsection{Emphatic Offline Actor-Critic (E-OAC)}
\label{app:eoac}


Because of the way emphatic traces are computed, the E-OAC agent requires learning from consecutive minibatches\footnote{Such that the first element of each rollout of a minibatch are adjacent to the last element of each rollouts of the previous minibatch}. Details can be found in Appendices \ref{app:oac} and \ref{app:eoac}. As explained in Appendix~\ref{app:sc2_interface}, we only apply policy improvement to the \texttt{function} and \texttt{delay} arguments of the action for simplicity.

The E-OAC agent uses the same BC agent as the OAC training in the previous section, that is, $\theta_0$ is also set to $\theta_V$. Then we run Algorithm~\ref{alg:eoac} with the same hyper-parameters as the OAC agent. However, unlike the OAC agent, it uses sequentially ordered trajectories in their order of interactions with the MDP, and reweight the policy gradient updates with the emphatic trace $F$. 

\begin{algorithm}
\caption{Emphatic Offline Actor-Critic (with fixed critic $V^\mu$)\label{alg:eoac}}
\begin{algorithmic}
\STATE \textbf{Inputs:} A dataset of trajectories $\mathcal{D}$, the mini batch size $M$, the rollout length $K$, an initial learning rate $\lambda$, the weights of the estimated behavior policy and value function from BC $\theta_0$ (such that $\hat\mu_{\theta_0}$ is the BC policy, and $V^{\hat\mu_{\theta_0}}$ is the behavior value function), the total number of observations processed $n_\mathrm{frames}$, the bootstrap length $N$, the IS threshold $\bar{\rho}$, a buffer $\cal{B}$ containing $M$ empty lists, a per-game step discount $\gamma_0$, initial emphatic traces $\forall j < N, F[j] = 1$.
\STATE {Set $\theta \leftarrow \theta_0$}
\FOR {$i=0..n_\mathrm{frames}/M-1$}
  \STATE{Set the gradient accumulator $g_\mathrm{acc} \leftarrow 0$.}
  \FOR {$j=0..M-1$}
    \IF {$\mathcal{B}[j]$ has less than $K+1$ elements}
      \STATE {Sample $T\sim\mathcal{D}$}
      \STATE {$\mathcal{B}[j] \leftarrow \mathrm{concatenate}(\mathcal{B}[j], T)$}
    \ENDIF
    \STATE {Set $X \leftarrow  \mathcal{B}[j][0:K+1]$}
    \STATE{Set $\mathcal{B}[j] \leftarrow \mathcal{B}[j][K:]$}
    \STATE{\emph{Compute the TD errors $\delta$, clipped IS ratios $\bar\rho$ and V-Trace targets $v$ with clipping threshold $\hat\rho$ as in Alg.~\ref{alg:oac}.}}
    \STATE{\emph{Compute the emphatic trace $F$:}}
    \FOR {$t=0..K-N-1$}
      \STATE{Set $F[t] = \prod_{p=1}^N (\gamma[t-p+1]\hat{\rho}[t-p]) F[t-N] + 1$}
    \ENDFOR
    \STATE{Set $g_t = \bar{\rho}[t] (X_r[t+1] + \gamma[t+1] v[t+1]  - V^{\hat\mu_{\theta_0}}(X_s[t]) \frac{\partial\log \pi_{\theta}(X_a[t]| X_s[t])}{\partial \theta}$}
    \STATE{Set $g_\mathrm{acc} \leftarrow g_\mathrm{acc} + \sum_{t=0}^{N-1} F[t] g_t$}
  \ENDFOR
  \STATE{Set $\lambda_i \leftarrow \frac{\lambda}{2}\left(\cos\left(\frac{i\pi}{n_\mathrm{frames}/M-1}\right) + 1 \right)$}
  \STATE{Set $\theta \leftarrow \theta - \lambda_i \cdot Adam(g_\mathrm{acc})$}
\ENDFOR
\RETURN{$\theta$}
\end{algorithmic}
\end{algorithm}

\subsubsection{MuZero (MZS and MZS-MCTS)}
\label{app:muzero}


We used \alphastar{}'s encoders and action prediction functions in the \mz{} architecture. \alphastar{}'s action prediction functions are notably autoregressive to handle the complex and combinatorial action space of \starcraft. We predict and embed the full action in the representation function, \emph{i.e.} for the root node in MCTS. To improve inference time, we only predict the \texttt{function} and \texttt{delay} arguments in the prediction heads in the model, \emph{i.e.} for non-root nodes in MCTS. We found that using as few as 20 actions for \mzsa{} worked well, and increasing it did not improve performance. 

We use a target network which is updated every $100$ learner steps, to compute the bootstrapped target value $v^{-}_{t+n}$. We found that $n=512$ worked best, which is notably large and roughly corresponds to half of the average game length.

Note that here we differ from \mz{} and \mzun{} by training with action and value targets obtained from the offline dataset. \mz{} and \mzun{} use the result of MCTS as the action and value targets.

We use Adam optimizer with additive weight decay \citep{loshchilov2017decoupled}, and a cosine learning rate schedule with $\lambda_0=10^{-3}$.

\begin{algorithm}
\caption{MuZero Supervised \label{alg:muzero}}
\begin{algorithmic} 
\STATE \textbf{Inputs:} A dataset of trajectories $\mathcal{D}$, the mini batch size $M$, the rollout length $K$, the temporal difference target distance $n$, an initial learning rate $\lambda$, the total number of observations processed $n_\mathrm{frames}$, and the initial weights used $\theta$ to parameterise the representation function $h_{\theta}$, the dynamics function $g_{\theta}$, and the prediction function $f_{\theta}$.
\FOR {$i=0..n_\mathrm{frames}/M-1$}
  \STATE{Set the gradient accumulator ${\nabla}_\mathrm{acc} \leftarrow 0$.}
  \FOR {$j=0..M-1$}
    \STATE {Sample a trajectory $T\sim\mathcal{D}$}
    \STATE {Sample $k$ in $[0, length(T))$}
    \STATE {Set $X \leftarrow T[k:k+K]$}
    \STATE {Set $X_\mathrm{TD} \leftarrow T[k + n]$}
    \STATE{Set ${\nabla}_\mathrm{acc} \leftarrow {\nabla}_\mathrm{acc} + \frac{1}{M} \cdot \frac{\partial \mathcal{L}_\mathrm{\mz{}}(\theta, X_s[k], X_a[k:k+K], X_\mathrm{TD})}{\partial \theta}$}.
  \ENDFOR
  \STATE{Set $\lambda_i \leftarrow \frac{\lambda}{2}\left(\cos\left(\frac{i\pi}{n_\mathrm{frames}/M-1}\right) + 1 \right)$}
  \STATE{Set $\theta \leftarrow \theta - \lambda_i \cdot Adam(g_\mathrm{acc})$.}
\ENDFOR
\RETURN{$\theta$}
\end{algorithmic}
\end{algorithm}

\subsection{Expanded win rate matrix}
\label{sec:expanded_win_matrix}

In this section, we present an expanded version of the win rate matrix of our reference agents shown in Figure~\ref{fig:win_matrix}. See Section~\ref{sec:reference_agents_evaluation} for more details.

\begin{figure}
    \centering
    \begin{subfigure}[t]{\textwidth}
        \centering
        \includegraphics[width=\textwidth]{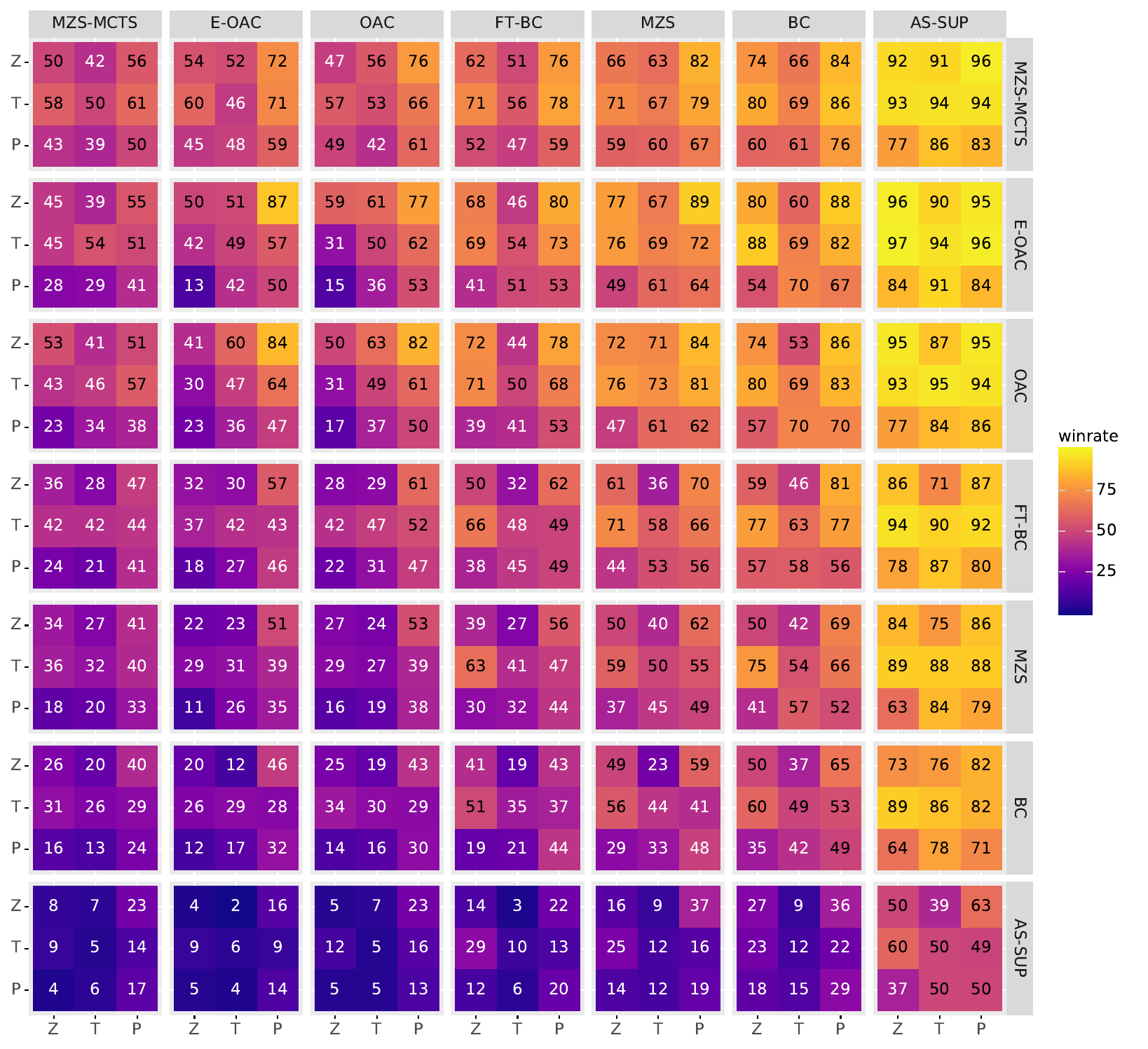}
    \end{subfigure}
    \caption{Win rate matrix of the reference agents broken down by race (Protoss, Terran, Zerg), normalized between $0$ and $100$. Note that because of draws, the win rates do not always sum to 100 across the diagonal. AS-SUP corresponds the the original AlphaStar Supervised agent trained to play all races.}
    \label{fig:expanded_win_matrix}
\end{figure}